\PassOptionsToPackage{table}{xcolor}
\documentclass[fleqn,10pt]{style/wlscirep}

\usepackage{hyperref}
\usepackage{url}
\usepackage{graphicx}
\usepackage{multirow}
\usepackage{booktabs}
\usepackage[utf8]{inputenc}
\usepackage{tcolorbox}
\usepackage{xcolor}
\usepackage[T1]{fontenc}    
\usepackage{booktabs}       
\usepackage{amsfonts}       
\usepackage{nicefrac}       
\usepackage{microtype}      
\usepackage{tikz}          
\usepackage{collcell}      
\usepackage{booktabs}
\usepackage{float}
\usepackage{fontawesome5}
\DeclareRobustCommand{\corrAuthor}{\textsuperscript{\faEnvelope[regular]}}
\usepackage{array}

\title{PathCowork: Learning from real-world pathology practice to build AI copilots}
\hypersetup{
    pdftitle={PathCowork: Learning from real-world pathology practice to build AI copilots},
    pdfsubject={Manuscript with Supplementary Information},
    pdfdisplaydoctitle=true
}
\author[1]{Sheng Wang}
\author[2]{Jialu Yao}
\author[1]{Maria V. Rossi}
\author[5]{Shunsuke Koga}
\author[1]{Esther Baranov}
\author[1]{MacLean P. Nasrallah}
\author[2]{Songhao Li}
\author[3]{Jonathan Xu}
\author[5]{Wei Du}
\author[1]{Cindy McGrath}
\author[1]{Timothy Chao}
\author[1]{Anne Wang}
\author[1]{Zhaohai Yang}
\author[1]{Daniel Akuma}
\author[1]{Selemon M. Walle}
\author[1]{Olga Pozdnyakova}
\author[1]{Edward B. Lee}
\author[1]{Zahra Alipour}
\author[1]{Annapurna Saksena}
\author[1]{Rashmi Tondon}
\author[1]{Nikolina Dioufa}
\author[1]{David E. Elder}
\author[1]{George J. Netto}
\author[1]{Xiaowei Xu}
\author[1,3,4,\corrAuthor]{Zhi Huang}
\affil[1]{Department of Pathology and Laboratory Medicine, Perelman School of Medicine, University of Pennsylvania, Philadelphia, PA, USA}
\affil[2]{Department of Electrical and Systems Engineering, School of Engineering and Applied Science, University of Pennsylvania, Philadelphia, PA, USA}
\affil[3]{Graduate Group in Genomics and Computational Biology, Perelman School of Medicine, University of Pennsylvania, Philadelphia, PA, USA}
\affil[4]{Department of Biostatistics, Epidemiology \& Informatics, Perelman School of Medicine, University of Pennsylvania, Philadelphia, PA, USA}
\affil[5]{Department of Pathology and Laboratory Medicine, Hospital of the University of Pennsylvania, Philadelphia, PA, USA}

\begin{abstract}
Medical AI learns from the answers doctors record, but rarely from how they interpret the evidence to reach them. Here we introduce PathCowork, a framework that turns pathologists' narrated examinations into supervision for visual grounding and procedures that guide diagnostic AI agents. Synchronized gaze, narration and navigation link visual findings to the process of interpreting them. Across 48 sessions from 16 pathologists, the platform captured 1,028 minutes of slide review. Using these data, we trained PathCowork-1T for server deployment and PathCowork-27B for local deployment, with improved visual grounding. Across five organ-specific evaluations, agents combining this grounding with recorded diagnostic procedures achieved primary task scores averaging 17.9 and 24.1 percentage points higher than Claude Code and Codex, respectively, with shorter processing times and lower inference costs. Compared to the state-of-the-art multiple-instance learning AIs, the PathCowork agent-proposed regions achieved higher prostate-grading scores with limited slide labels. PathCowork can also help construct pathology AI benchmarks where questions derived from the recordings identified difficulties in recognizing viewing context and rejecting inapplicable premises. In summary, PathCowork proposes a new paradigm to scale up the data collection and train pathology AI through transparent reasoning from diagnostic sessions, opening a path towards systems that learn directly from routine clinical practice.
\end{abstract}
 
\begin{document}
\raggedbottom
\pdfbookmark[0]{PathCowork}{paper.title}
\maketitle
\vspace{-5em}
{\noindent\corrAuthor~To whom the correspondence should be addressed: Zhi Huang (\href{mailto:zhi.huang@pennmedicine.upenn.edu}{zhi.huang@pennmedicine.upenn.edu})\par}
\vspace{3em}
\section*{Abstract}
{\noindent\theabstract\par}
\vspace{8pt}

\section*{Main}
\pdfbookmark[1]{Main}{paper.main}

Rising concerns, as well as regulatory frameworks, increasingly emphasize clinicians' ability to understand AI outputs. For instance, the FDA requires radiological computer-assisted diagnosis software to include labeling that explains how users should interpret its outputs~\cite{fda2025cadinterpretation}. Concurrently, pathology AI commonly learns from images paired with diagnostic reports to classify disease and predict outcomes~\cite{ding2025titan,shaikovski2024prism}.
However, these written reports preserve the outcome of interpretation while leaving much of the ``reasoning'' process itself unrecorded.
In pathology, the reasoning behind the diagnosis plays a critical role: a doctor may survey the tissue, inspect suspicious regions, weigh alternative diagnoses and revisit uncertain findings before reaching a conclusion.
The final report rarely captures the complexity of interpretation behind these decisions.
Scaling up collections of images and reports cannot recover the ``thinking'' process that was never recorded.
In contrast, pathologists get educated to diagnose tissue through guided examination and discussion with experienced colleagues~\cite{crowley2003development,drew2021scanning}. At a shared microscope, trainees follow an expert's search through the tissue and hear the reasoning behind each decision. They learn to connect what they see with why it matters and what to examine next. However, this experience remains largely outside how medical AI has been trained contemporarily: they were taught what doctors concluded, but not how they interpreted the evidence. As a consequence, AI models learn to minimize the training loss rather than to emulate the reasoning process that makes a diagnosis credible and interpretable to humans.

Contemporary pathology AI typically learns slide-level predictions from tile features, either through multiple-instance learning (MIL), which aggregates tiles using only slide labels~\cite{ilse2018attention,campanella2019clinical,lu2021clam}, or through transformer-based slide encoders built on pretrained tile encoders~\cite{ding2025titan,shaikovski2024prism}, to address diagnostic and prognostic tasks.
These approaches rely on increasingly powerful foundation models, including vision-only encoders such as UNI and Virchow~\cite{chen2024uni,vorontsov2024virchow} and vision--language models such as PLIP~\cite{huang2023plip}, CONCH~\cite{lu2024conch} and MUSK~\cite{xiang2025musk}. More recently, PathChat and SlideChat have coupled pathology vision encoders with large language models to support diagnostic dialogue~\cite{lu2024pathchat,chen2025slidechat}.
Despite their differences, these models share a common form of aforementioned ``weak'' supervision: they learn from images alone or from images paired with labels, captions or reports, without access to the examination process through which pathologists reach a diagnosis.
Each approach therefore leaves part of this process unaddressed. In MIL, the attention that drives slide-level predictions is learned indirectly from slide labels and need not correspond to the regions a pathologist would examine. PathChat supports diagnostic conversation but offers limited localization of the evidence behind its answers~\cite{lu2024pathchat}. Most of these models also offer clinicians limited means to adapt them to their own diagnostic practice. 
General-purpose coding agents such as Codex and Claude Code can navigate slides and expose their intermediate steps, but their underlying models are not trained on pathologists' diagnostic experience and are difficult to deploy within institutional infrastructure.

This demonstrates an urgent dual challenges and unmet needs: \textbf{First}, the expertise expressed in everyday clinical practice remains largely disconnected and unrecorded that could be leveraged to train medical AI; and \textbf{Second}, there is no such pathology AI to train the actual expert reasoning for a more tranparent and reliable copilot. Work on selective navigation and expert gaze has begun to bring viewing behavior into model training~\cite{dong2018razn,xiong2023hagmil,buckley2025giant,xu2026mmnavagent}. Our previous Pathology-CoT study showed that recorded viewing behavior paired with verified reasoning could supervise agents navigating tissue~\cite{wang2026pathologycot}. Motivated by such unmet need and the existing gap in the field, here we introduce PathCowork, which represents a fundamental paradigm shift in how pathology data is recorded and how medical AI learns, from training the diagnostic conclusions to the training of the actual practice of interpretation (\textbf{Figure~\ref{fig:main}}). For the \textbf{data collection}, PathCowork captures diagnostic examinations as they unfold within a whole slide viewer ``TissueLab''~\cite{li2025co}, linking pathologists' observations and explanations to the tissue they are examining. Across 48 sessions, 16 pathologists contributed 1,028 minutes of narrated slide review, with gaze and navigation recorded alongside their observations.
These records allow us to \textbf{develop powerful agentic AI frameworks} to locate relevant features and provide procedures that guide agents through new examinations, making their evidence and diagnostic steps transparent for human audibility (\href{https://github.com/zhihuanglab/PathCowork/blob/main/videos/01_Main_Video_1_Agent.mp4}{\textbf{Main Video 1}}).
This is established through finetuning open-source vision-language models (e.g., Kimi and Qwen)~\cite{moonshot2026kimik26,yang2025qwen3} on these records to learn visual grounding and used the recorded observations to construct diagnostic procedures.

In our benchmarks, fine-tuning improved aggregate spatial grounding accuracy, which measures the model's ability to precisely locate specified pathological features within images, from 5.6\% to 52.2\%, compared with 24.4\% for a frontier model at the time of evaluation.
Across five organ-specific evaluations, agents combining the learned grounder with recorded examination procedures achieved primary task scores averaging 17.9\% and 24.1\% higher than Claude Code and Codex, respectively.
Averaged across the same tasks, inference costs were 11.0- and 10.3-fold lower, with 6.2- and 4.7-fold speedups.
PathCowork agent also supplied regions for slide-level MIL training, improving prostate grading when labelled training slides were limited.
Additionally, PathCowork converts recorded examinations into case-specific benchmark questions that can test whether a model can use the visual context of a review, including recognizing the viewing scale and tissue location, rejecting questions with inapplicable premises, integrating evidence across slides and choosing where to look next.
These questions depended strongly on the images: on the same 148 questions, Claude Opus 4.8 answered 26.4\% correctly from text alone and 81.8\% with images, a 55.4-percentage-point difference, compared with 59\% and 74\% on PathMMU.
Compared with a pathology resident, the evaluated models showed the largest gaps in recognizing the viewing context and rejecting inapplicable premises, which implicates that the frontier models still struggle to recognize what they are viewing and when a question does not apply, and underscoring the need for (i) more rigorous benchmarks for pathology; and (ii) pathology-specific vision--language ``reasoning'' models.

Because PathCowork learns from the same process of observation and explanation through which pathologists are trained, routine diagnostic practice could become a continuing source for AI to learn passively. Each examination could contribute to a growing body of experience that informs successive generations of medical AI. Two fine-tuned open-source models, PathCowork-1T for server deployment and PathCowork-27B for local deployment, are released at \url{https://github.com/zhihuanglab/PathCowork}, making the learned grounding available for reuse with diagnostic procedures.
 
\clearpage
\section*{Results}
\pdfbookmark[1]{Results}{paper.results}

\subsection*{Capturing the experience behind histopathological diagnoses}

A whole-slide image and its diagnostic report describe the tissue and the conclusion, but leave much of the examination unrecorded: where a feature was found, the scale at which it became visible and why the pathologist examined it next.
We developed PathCowork to capture these intermediate observations during slide review.
Within a desktop-based whole-slide image viewer ``TissueLab''~\cite{li2025co}, pathologists examined cases while explaining their findings as they would to a trainee.
A screen-mounted eye tracker and microphone recorded gaze and narration alongside mouse movements, viewport position and magnification (\textbf{Figure~\ref{fig:main}a,b,c}).
An example of these synchronized observations during skin-biopsy review is shown in \href{https://github.com/zhihuanglab/PathCowork/blob/main/videos/02_Video_2_Behavior_Acquisition.mp4}{\textbf{Main Video 2}}.
We processed these recordings after each session to construct training examples.

The platform connected users, viewing environments and processing models (\textbf{Figure~\ref{fig:main}d}), producing observations that linked narration to the corresponding visual context (\textbf{Figure~\ref{fig:main}e}).
The resulting records retained the image region, description, viewing trajectory and revisits, including uncertainty in the interpretation.
For example, a description of peripheral palisading and myxoid stroma could be revisited alongside the corresponding tissue, whereas a question about an infiltrative tumor border remained a tentative observation (\textbf{Figure~\ref{fig:main}f}).
Viewing-event processing, speech alignment, fixation filtering and cross-modal review produced these linked records (\textbf{Figure~\ref{fig:main}g}; \textbf{Methods}), supplying localized examples and explanations without a separate drawing or annotation step during the recorded examination.

The collection comprised 48 sessions and 1,028 minutes of review from 16 pathologists: six attending pathologists, four fellows and six residents (\textbf{Figure~\ref{fig:main}h}).
Skin accounted for 423 minutes, followed by colon and lymph-node review (245 minutes), diffuse large B-cell lymphoma (102 minutes) and glioblastoma (99 minutes), with additional breast, prostate, cytology, bladder, neurodegenerative disease and stomach recordings.
Among 4,759 categorized sentences, narration included diagnosis, architecture, cellular morphology, navigation and expressions of uncertainty (\textbf{Figure~\ref{fig:main}i}).
Vocabulary also reflected the tissue being examined. For example, all five readers in the skin analysis used ``nest'' and ``superficial'' more often in skin than in the pooled non-skin records, with log-odds $z$-scores of 1.7--3.6 and 0.7--3.3, respectively (\textbf{Figure~\ref{fig:main}j}).

Compared with WSI--report pairs, these records retain the location of an observation and the ``reasoning'' that led to it (\textbf{Figure~\ref{fig:main}k,l}). They support two uses developed below: training AI models to identify visual evidence and assembling diagnostic procedures that can be inspected and reused as guidelines.
The system comparison in \textbf{Figure~\ref{fig:main}m} places these capabilities alongside adaptation to a clinician's practice and deployment within local infrastructure.
It describes what each system makes available; the subsequent experiments evaluate grounding and task performance.

\newgeometry{left=1.5cm,right=1.5cm,top=1.0cm,bottom=1.0cm}
\begin{figure}[p]
    \pdfbookmark[2]{Figure 1}{paper.figure.1}
    \centering
    \includegraphics[width=180mm,height=\dimexpr\textheight-32pt\relax,keepaspectratio]{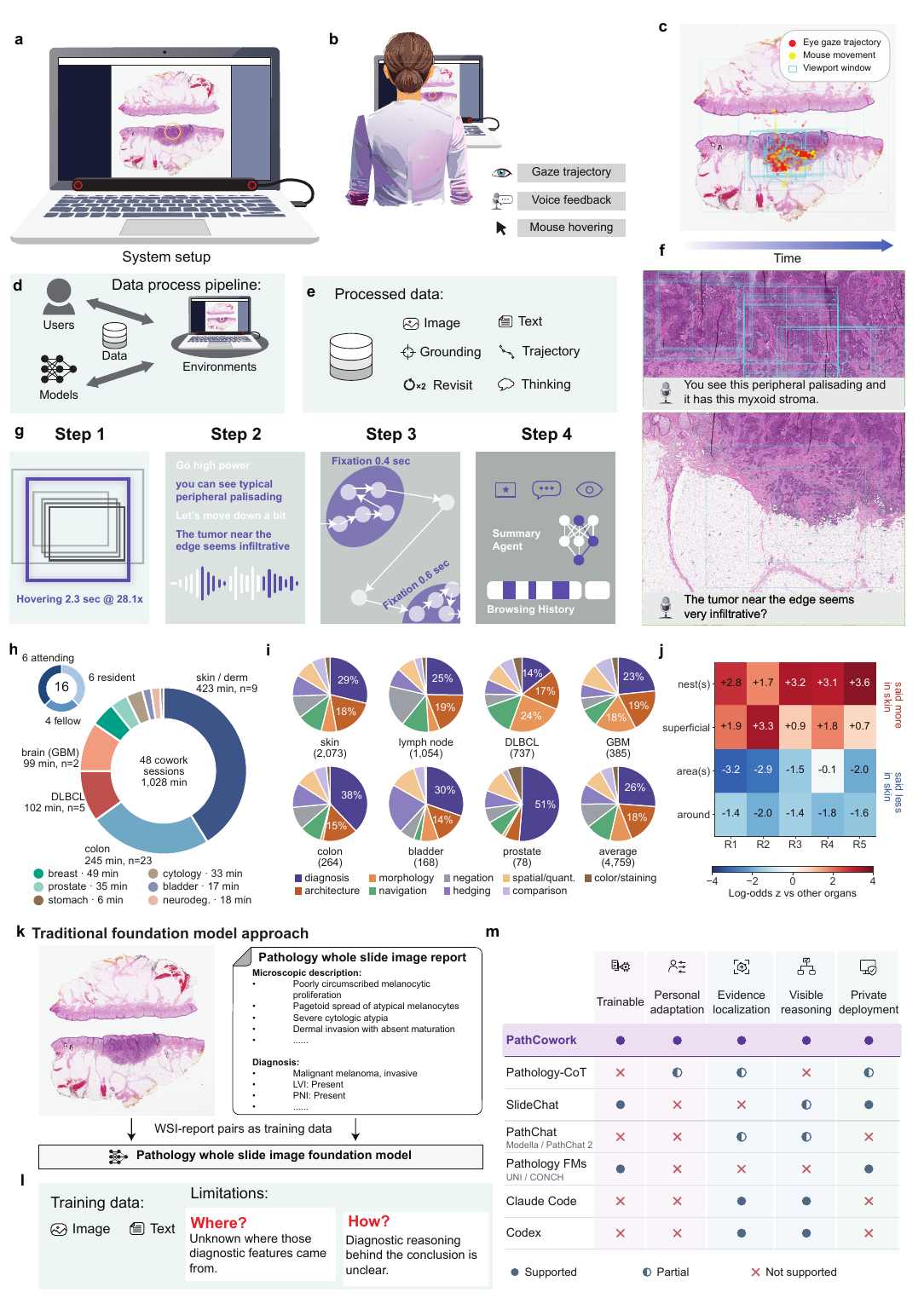}
    \caption{Legend on page~\pageref{cap:fig:main}.}
    \label{fig:main}
\end{figure}
\restoregeometry
\begin{figure}[H]
    \ContinuedFloat
    \caption{\textbf{Recording and organizing the experience of slide review.}
    \textbf{a,} Recording setup with a whole-slide image (WSI) viewer, eye tracker and microphone.
    \textbf{b,c,} Synchronized gaze, narration, mouse movements and viewport history during examination.
    \textbf{d,} System infrastructure connecting users, models, data and viewing environments.
    \textbf{e,} Structured observations retaining images, text, grounding, trajectories, revisits and interpretations.
    \textbf{f,} Examples of observations linked to the tissue being examined, including peripheral palisading, myxoid stroma and uncertainty about an infiltrative border.
    \textbf{g,} Processing of viewing events, speech alignment, gaze fixations and review history.
    \textbf{h,} Recording duration by tissue or disease group and participant training levels. The current collection contains 48 sessions, 1,028 min and 16 pathologists; $n$ denotes sessions.
    \textbf{i,} Narration categories by tissue or disease group and in the pooled set; numbers in parentheses denote categorized sentences. Each sentence received one primary LLM-assigned category; 2,879 sentences assigned to ``other'' were excluded, leaving 4,759. Each pie has equal area; diagnosis, architecture and cellular morphology are annotated when they account for at least 10\% of sentences.
    \textbf{j,} Vocabulary enrichment in each of five skin readers relative to the pooled non-skin records. Cells show log-odds $z$-scores from segment-level lexeme presence with a 0.5 cell correction; positive and negative values indicate higher and lower usage, respectively. Terms were selected exploratorily.
    \textbf{k,l,} WSI--report pairs and the spatial and procedural information not specified by those pairs.
    \textbf{m,} Comparison of seven systems or model families across trainability, personal adaptation, evidence localization, visible reasoning and private deployment. Filled circles indicate support, half-filled circles partial support and crosses lack of availability in the compared configuration (Methods). Visible reasoning denotes inspectable examination steps and associated evidence. PathChat has partial support for evidence localization; the representative pathology foundation models (UNI and CONCH) do not provide this output as standalone encoders. PathCowork supports all five capabilities, with model weights released.}
    \label{cap:fig:main}
    \caption*{Artwork on page~\pageref{fig:main}.}
\end{figure}

\subsection*{Learning visual grounding from narrated examination}

The recordings linked diagnostic language to the scale at which a finding was examined.
When mitosis was mentioned, the median viewport magnification was $23\times$ (19 observations), compared with $21\times$ for nuclei and $13\times$ for cytoplasm (\textbf{Figure~\ref{fig:behavior}a}).
Anatomical structures were generally discussed at lower power: median magnification was $4.0\times$ for glands, $3.0\times$ for dermis, $2.6\times$ for follicles and $1.4\times$ for lymph nodes (103 observations).
Although the distributions overlapped, the median magnification for mitosis was approximately 16 times that for lymph nodes, linking each concept to the scale at which pathologists examined it.

Spatial descriptions were associated with different extents of gaze movement (\textbf{Figure~\ref{fig:behavior}b}).
The median root-mean-square spread of gaze around an utterance was 3.0\,mm for ``diffuse'' and 2.3\,mm for ``infiltrate'', compared with 1.7\,mm for ``around'', 1.5\,mm for ``sheets'', 1.1\,mm for ``nests'' and 0.7\,mm for ``scattered''.
These measurements describe the area explored around the spoken observation.
Together, the viewing scale and spatial context provide information that a transcript alone does not specify.

We used observations linked to their corresponding regions in the original slides to supervise \textbf{visual grounding}, a critical task in pathology in which a model is given an image and a specified concept and must localize the region corresponding to that concept.
The examples in \textbf{Figure~\ref{fig:behavior}c} show how narrated microvascular proliferation and mitosis provide training targets for locating those findings.
Supervised fine-tuning improved aggregate grounding in both 1T- and 27B-parameter open-source models (Kimi K2.6 and Qwen 3.6 27B)~\cite{moonshot2026kimik26,yang2025qwen3} relative to their unmodified counterparts (\textbf{Figure~\ref{fig:behavior}d}).
The score rose from 19.8\% to 51.2\% for the 1T model and from 5.6\% to 52.2\% for the 27B model, gains of 31.4 and 46.6 percentage points, respectively.
Both adapted models exceeded the GPT-5.4 reference of 24.4\%.

We examined this adaptation more closely using seven melanocytic concepts, with training examples drawn from dermatopathology recordings pooled across participating pathologists.
Concept recall, precision and spatial grounding were evaluated separately against Claude Opus 4.7, GPT-5.4 and Gemini 3.1 (\textbf{Figure~\ref{fig:behavior}e}).
The adapted localizer showed stronger grounding for the dermoepidermal junction, dermal melanocytic nests and pagetoid spread, although recognizing a finding did not ensure accurate localization.
The qualitative examples make this distinction visible: PathCowork followed the dermoepidermal junction and marked a localized epidermal region for pagetoid spread, whereas comparison outputs included contours displaced above the tissue or regions covering much of the field (\textbf{Figure~\ref{fig:behavior}f}).
Additional examples are shown in \textbf{Supplementary Figures~\ref{fig:skin_grounding_1} and~\ref{fig:skin_grounding_2}} (Supplementary Note~\ref{supp:training}).

\clearpage
\newgeometry{left=1.5cm,right=1.5cm,top=1.0cm,bottom=1.0cm}
\begin{figure}[p]
    \pdfbookmark[2]{Figure 2}{paper.figure.2}
    \centering
    \includegraphics[width=180mm,height=\dimexpr\textheight-32pt\relax,keepaspectratio]{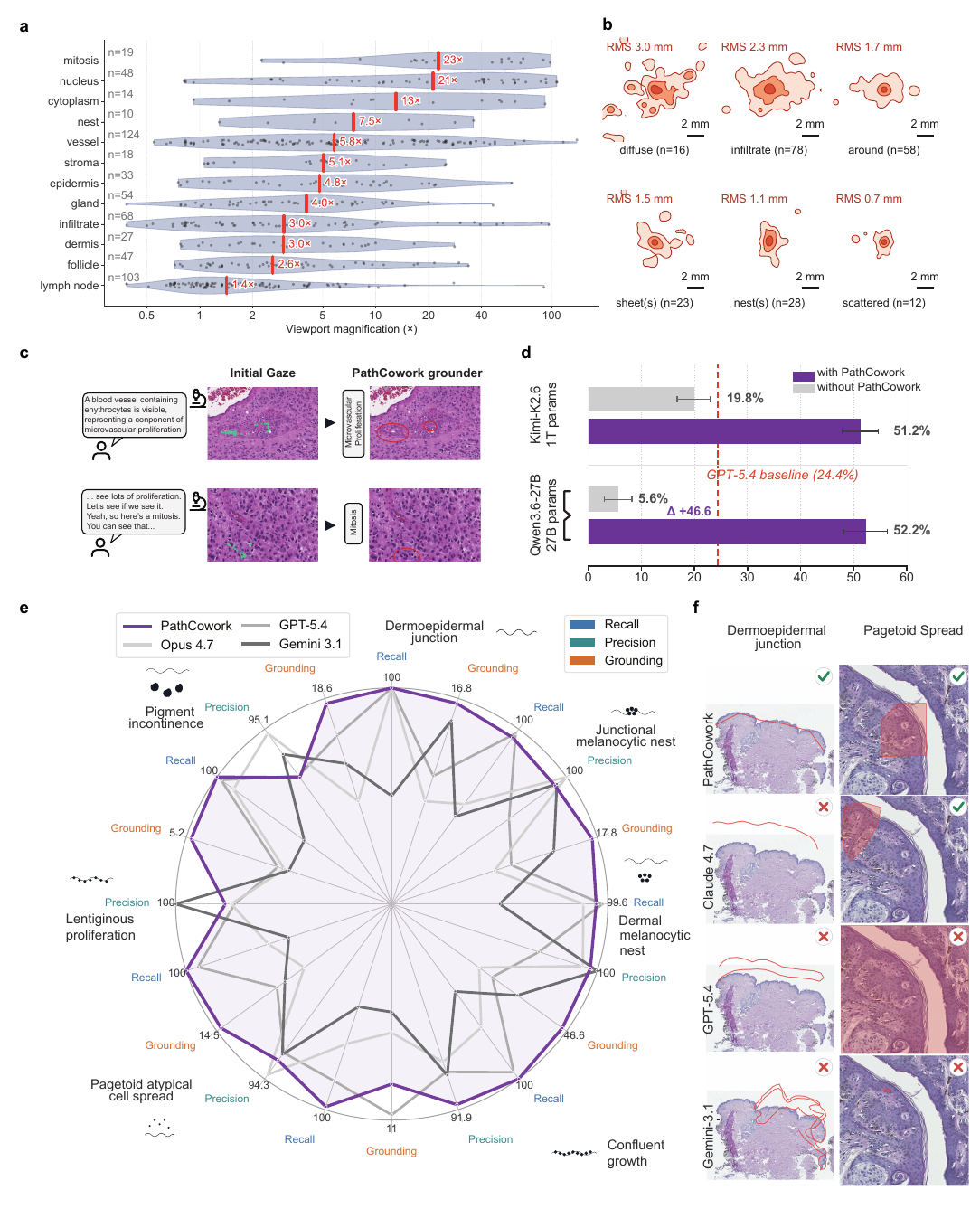}
    \caption{Legend on page~\pageref{cap:fig:behavior}.}
    \label{fig:behavior}
\end{figure}
\restoregeometry
\begin{figure}[H]
    \ContinuedFloat
    \caption{\textbf{Learning visual grounding from recorded behavior.}
    \textbf{a,} Viewport magnification associated with mentions of diagnostic concepts. Dots represent observations, violin widths indicate density, and red lines and labels show medians; $n$ denotes concept-associated segments. Magnification is the objective-equivalent viewport scale from per-slide optics metadata; unreliable rotated viewports were excluded.
    \textbf{b,} Gaze-density contours associated with six spatial terms, aligned on each utterance's gaze centroid and pooled by term. Annotations give the median per-utterance root-mean-square (RMS) gaze spread. Gaze was pooled from segments overlapping the utterance and its surrounding 5 s, retaining windows with at least three samples. Nested contours enclose 20\%, 50\% and 80\% of smoothed in-field gaze density from dark to light; scale bars, 2 mm.
    \textbf{c,} Narrated microvascular proliferation and mitosis, showing recorded gaze and the corresponding findings located by the PathCowork grounder.
    \textbf{d,} Aggregate grounding performance of Kimi-K2.6 (1T parameters) and Qwen3.6-27B after supervised fine-tuning on PathCowork recordings (purple) and without fine-tuning (grey). Scoring uses Dice on filled polygon masks or equal-width, 50-pixel bands around predicted and reference polylines. Bars show the mean of the organ-level average Dice scores; error bars show the standard deviation across organs. The dashed line denotes GPT-5.4; the annotated difference for the 27B model is in percentage points.
    \textbf{e,} Concept recall, precision and spatial grounding for the PathCowork-1T localizer, Claude Opus 4.7, GPT-5.4 and Gemini 3.1 across seven melanocytic concepts, with 50 test image--concept pairs per concept (350 pairs in total). Test patients were separate from training patients. Radar axes are normalized separately; perimeter values label the individual scales. Training used pooled dermatopathology recordings. Missed targets and unparseable spatial responses receive zero localization scores and remain in the evaluation denominator.
    \textbf{f,} Qualitative localization of the dermoepidermal junction and pagetoid spread by the four models, shown as outlines and shaded regions. Ticks and crosses indicate correct and incorrect localization, respectively.}
    \label{cap:fig:behavior}
    \caption*{Artwork on page~\pageref{fig:behavior}.}
\end{figure}

\subsection*{PathCowork compiles diagnostic experience into AI agents}

As the grounded observations describe what to look for, the accompanying narration also describes how to proceed through an examination.
PathCowork uses a diagnostic task to retrieve relevant review histories and organize their observations into a reusable diagnostic procedure (\textbf{Figure~\ref{fig:skill}a,b,c}).
For melanoma versus dysplastic nevus, the retrieved material includes examination of the dermoepidermal junction, assessment of junctional growth and the dermal component, and a search for mitoses at higher power.
The resulting skill links these steps to the anatomical targets, viewing scales and visual concepts needed to carry them out.
Viewer logs or gaze associate the narrated concepts with tissue regions, providing paired images and spatial targets for grounding (\textbf{Figure~\ref{fig:skill}c}).
These targets can trace a boundary, such as the dermoepidermal junction, or outline regions, such as dermal melanocytic nests.
The agent can then use the learned grounder to locate those concepts on another slide.
This construction separates the order and purpose of examination from the model that supplies its visual localization, while retaining their common source in recorded cases.

We selected five diagnostic tasks to address common challenges in cancer diagnosis and draw on the expertise of our collaborating pathologists. The resulting skills covered melanoma versus dysplastic nevus, prostate needle-biopsy assessment, glioblastoma versus lower-grade glioma, colon lymph-node metastasis detection and Nottingham grading of invasive breast carcinoma (\textbf{Figure~\ref{fig:skill}d}).
Each procedure specified a sequence appropriate to its task.
The lymph-node skill, for example, first inventories the tissue and identifies node boundaries, then directs examination of the capsule, sinuses and node interior before zooming into a detected focus.
The breast skill instead organizes examination around tubule formation, nuclear pleomorphism and a mitotic hotspot.

We evaluated these agents alongside Claude Code and Codex using the primary outcome for each task (\textbf{Figure~\ref{fig:skill}e}).
Giving each of the five tasks equal weight, primary scores were on average 17.9 percentage points higher for PathCowork than Claude Code and 24.1 points higher than Codex.
For melanoma versus dysplastic nevus, PathCowork achieved 80\% accuracy across 20 cases comprising 69 whole-slide images, compared with 65\% for each comparator.
It detected all melanomas in this evaluation, with 60\% specificity, whereas both comparators had 40\% melanoma sensitivity and 90\% specificity (\textbf{Supplementary Figure~\ref{fig:organ_supp}}; Supplementary Note~\ref{supp:agents}).
The higher accuracy therefore involved a trade-off between missing melanoma and overcalling dysplastic nevi.
For prostate assessment, the cancer-detection score was 95\%, compared with 85\% (17/20) for Claude Code and 35\% for Codex.
On the same 30 brain cases, PathCowork achieved 86.7\% accuracy (26/30), compared with 60\% (18/30) for Claude Code and 70\% (21/30) for Codex. PathCowork correctly classified 13/15 cases in each class; Claude Code and Codex correctly classified 3/15 and 6/15 GBM cases, respectively, and all 15 LGG cases.
Lymph-node detection recall was 97\%, 87.3\% and 86\%, respectively, and breast grade accuracy was 71.8\% (28/39), 43.6\% (17/39) and 53.8\% (21/39), using the same 39 cases for all three agents. Component-score accuracy and grade-specific recall show where these differences arise (\textbf{Supplementary Figure~\ref{fig:organ_supp}}).

The higher task scores were accompanied by shorter processing times across all five tasks, with 6.2- and 4.7-fold speedups over Claude Code and Codex, respectively, calculated from the task-averaged times (\textbf{Figure~\ref{fig:skill}e}).
For skin, PathCowork required a mean of 39\,s per slide, compared with 252\,s for Claude Code and 117\,s for Codex.
For brain classification, the corresponding times were 20\,s, 332\,s and 285\,s; for breast grading, they were 106\,s, 300\,s and 192\,s.
Colon lymph-node examination took a median of 22\,s per slide with PathCowork, compared with 277\,s and 322\,s for the two comparators.

Reported inference costs were also lower for PathCowork in every task, ranging from \$0.01 to \$0.17 per slide, compared with \$0.47--\$1.98 for Claude Code and \$0.75--\$1.25 for Codex.
The ratio of the task-averaged costs corresponded to 11.0- and 10.3-fold reductions, respectively.
Skin examination cost \$0.17 per slide with PathCowork, versus \$1.00 and \$0.75, respectively; colon lymph-node examination cost \$0.01, versus \$0.96 and \$1.04.
These comparisons evaluate the complete behavior-derived agents: recordings supply the procedural skills, their execution sequences and the supervision used to adapt the grounder.

Model weights for \textbf{PathCowork-1T} (server deployment) and \textbf{PathCowork-27B} (local deployment) are released at \url{https://github.com/zhihuanglab/PathCowork}.
Both learn visual grounding from the recorded observations and can supply the localization used by the diagnostic procedures.
The agent results above use the PathCowork-1T grounder; the paired grounding comparison evaluates both models (\textbf{Figure~\ref{fig:behavior}d}).
To further leverage the full breadth of pathologist behavior collected across multiple organs and diseases, beyond the five diseases evaluated in our diagnostic benchmarks, we explored \emph{General navigation learned from pathologist behavior}, implemented as PathCowork-general, which learns generalizable viewing actions to support evidence gathering and subsequent LLM interpretation for open-ended diagnostic questions not yet covered by a distilled skill (Supplementary Note~\ref{supp:so1}; \textbf{Supplementary Figure~\ref{fig:so1_navigator}}; \href{https://github.com/zhihuanglab/PathCowork/blob/main/videos/03_Supplementary_Video_3_General_Navigation.mp4}{\textbf{Supplementary Video 3}}).

\clearpage
\newgeometry{left=1.5cm,right=1.5cm,top=1.0cm,bottom=1.0cm}
\begin{figure}[p]
    \pdfbookmark[2]{Figure 3}{paper.figure.3}
    \centering
    \includegraphics[width=180mm,height=\dimexpr\textheight-32pt\relax,keepaspectratio]{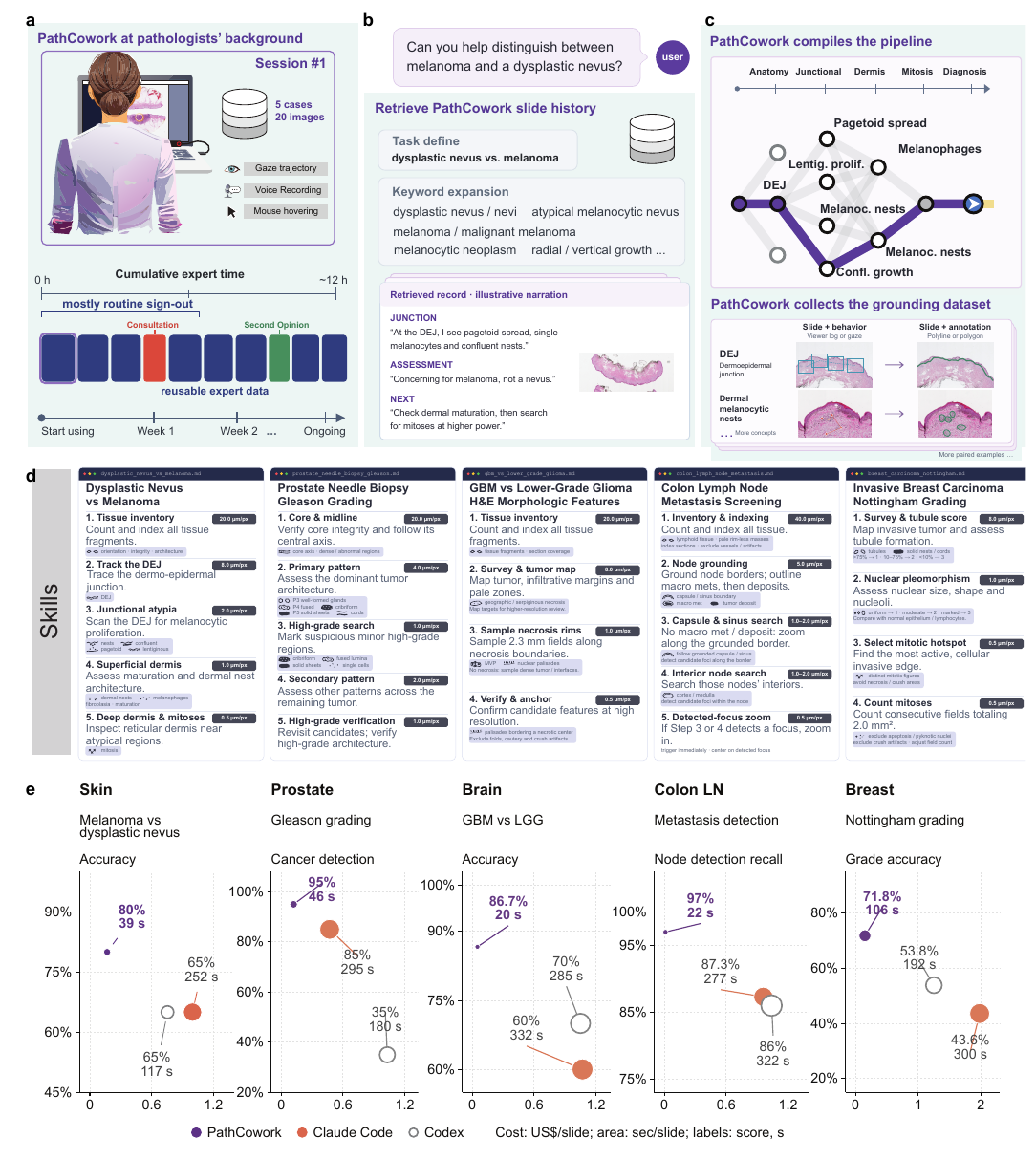}
    \caption{Legend on page~\pageref{cap:fig:skill}.}
    \label{fig:skill}
\end{figure}
\restoregeometry
\begin{figure}[H]
    \ContinuedFloat
    \caption{\textbf{Constructing and evaluating diagnostic agents from recorded experience.}
    \textbf{a,} Collection of multimodal review histories across sessions.
    \textbf{b,} Retrieval of relevant histories using a diagnostic task and expanded queries, with an illustrative narrated record.
    \textbf{c,} Compilation of diagnostic steps and collection of visual examples for grounding. The branching schematic represents the examination procedure. Below, paired images illustrate behavioral context from viewer logs or gaze (left) and corresponding spatial targets (right): a polyline for the dermoepidermal junction and polygons for dermal melanocytic nests. Overlays are schematic; stacked cards and ellipses indicate additional examples and concepts.
    \textbf{d,} Five compiled skills for skin, prostate, brain, colon lymph nodes and breast, in that order. The excerpts specify examination steps, targets and viewing scales.
    \textbf{e,} Task performance versus cost per slide for PathCowork (purple), Claude Code (orange) and Codex (grey outline). Bubble area represents time per slide; labels give the score and time. Primary metrics are skin diagnostic accuracy, prostate cancer detection, brain diagnostic accuracy, colon node detection recall and breast grade accuracy. Colon time is the median; other displayed times are means. Task-specific cohorts and configurations are detailed in Supplementary Note~\ref{supp:agents}.}
    \label{cap:fig:skill}
    \caption*{Artwork on page~\pageref{fig:skill}.}
\end{figure}

\subsection*{PathCowork guides slide-level learning with behavior-derived regions}

The adapted agent could also supply spatial guidance for training another model.
We tested this with our proposed ``CoworkMIL'', which uses regions proposed by PathCowork to guide multiple-instance learning (MIL) for prostate grading (\textbf{Figure~\ref{fig:coworkmil}a}).
The agent examined slides in the MIL training set and identified regions of interest.
These regions supplied tile-level supervision alongside the slide labels, directing training toward the tissue selected by the grounder.
The regions were required during training only; the trained model subsequently predicted grades from the slide images.

We compared CoworkMIL with attention-based MIL (ABMIL)~\cite{ilse2018attention} and a CLAM-inspired pseudo-label control~\cite{lu2021clam}, using three representative pathology image encoders (UNI2-h, Virchow2 and CONCH), 10--50 training slides and 20 seeds per setting (\textbf{Figure~\ref{fig:coworkmil}b,c,d,e,f}).
At 20 training slides, CoworkMIL increased mean quadratic-weighted $\kappa$ over ABMIL from 0.466 to 0.544 with UNI2-h, from 0.385 to 0.486 with Virchow2, and from 0.484 to 0.585 with CONCH.
It also exceeded the mean performance of the pseudo-label control with each encoder.
The largest gain over ABMIL was 0.151 with Virchow2 at 30 slides ($\kappa=0.620$ versus 0.469).
The benefit depended on the amount of training data and the encoder: at 50 slides, the gap over ABMIL narrowed to 0.013 with Virchow2 and became negative with UNI2-h, whereas CONCH retained an advantage ($\kappa=0.741$ versus 0.643).
Across the tested training-set sizes, gains were most consistent with CONCH.

In the auxiliary-loss sweep, $\kappa$ increased from 0.500 at weight $\beta=0.1$ to 0.583 at $\beta=1.0$, then changed little at $\beta=2.0$ (0.576; \textbf{Figure~\ref{fig:coworkmil}g}).
In the illustrated biopsy, CoworkMIL concentrated attention on glandular tissue, whereas the highest-attention ABMIL patches included more surrounding stroma (\textbf{Figure~\ref{fig:coworkmil}h}).

As a separate comparison, EAGLE~\cite{neidlinger2026eagle}, which selects tiles with frozen attention and fits a slide-level linear classifier, achieved mean $\kappa$ of 0.430 and 0.435 at 0.5 and 2.0\,$\mu$m per pixel, respectively, using the same 20-slide training subsets and 30-slide test cohort (Supplementary Note~\ref{supp:eagle}). These values were below the 0.486 obtained with Virchow2-CoworkMIL, although the pipelines differ in spatial sampling, feature representation and training supervision.

\clearpage
\newgeometry{left=1.5cm,right=1.5cm,top=1.0cm,bottom=1.0cm}
\begin{figure}[p]
    \pdfbookmark[2]{Figure 4}{paper.figure.4}
    \centering
    \includegraphics[width=180mm,height=\dimexpr\textheight-32pt\relax,keepaspectratio]{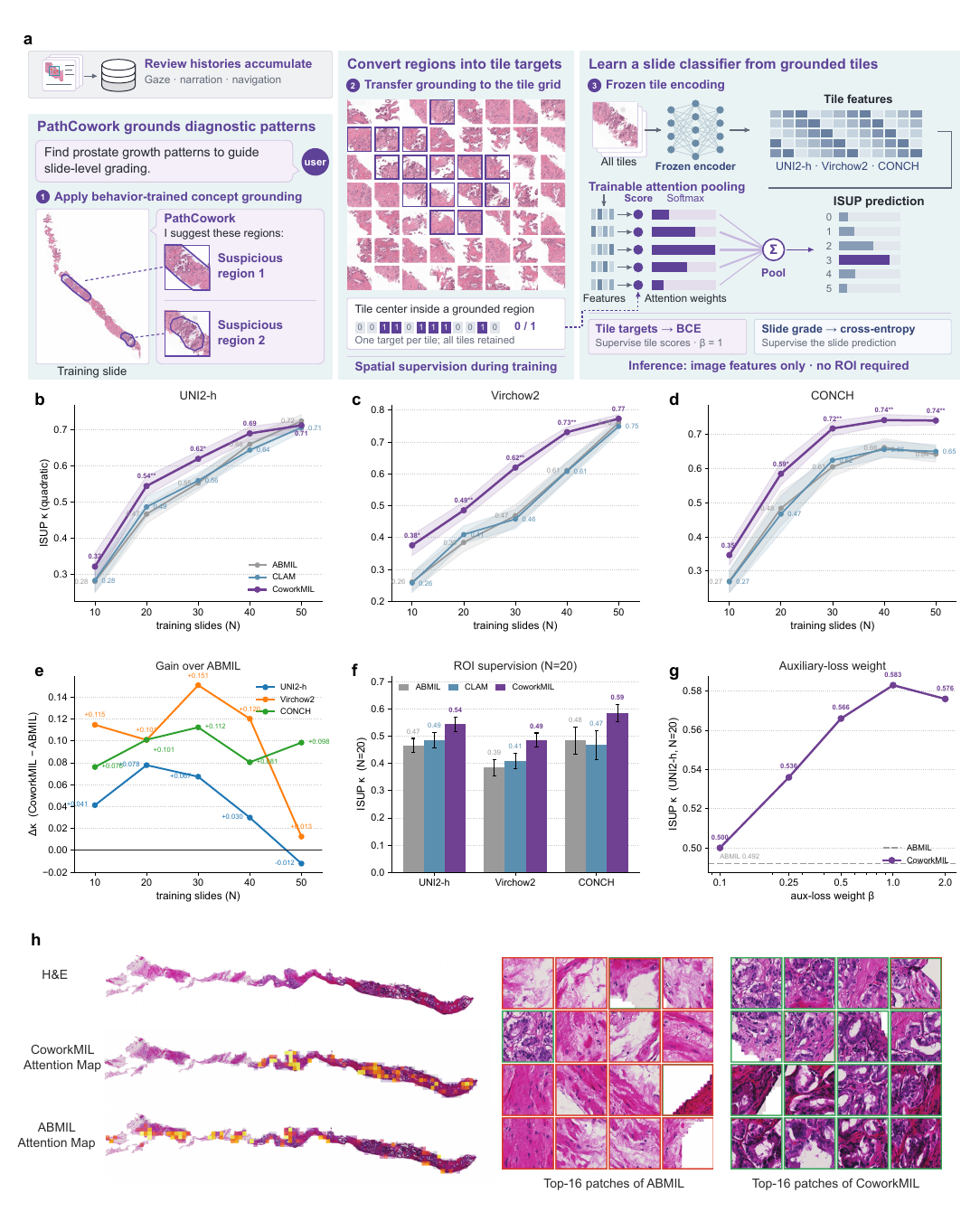}
    \caption{Legend on page~\pageref{cap:fig:coworkmil}.}
    \label{fig:coworkmil}
\end{figure}
\restoregeometry
\begin{figure}[H]
    \ContinuedFloat
    \caption{\textbf{PathCowork-guided multiple-instance learning for prostate grading.}
    \textbf{a,} Expert review histories provide behavioral supervision for the PathCowork grounder, which proposes suspicious regions on multiple-instance learning (MIL) training slides. Region polygons are converted into binary tile targets. A frozen image encoder supplies tile features; learned tile logits receive auxiliary binary supervision and define attention weights for slide-level pooling. Training combines slide-label cross-entropy with tile-level binary cross-entropy, with a fixed auxiliary weight of $\beta=1$ in \textbf{b,c,d,e,f}. PathCowork guidance is used only during training; inference requires the slide alone. Network glyphs and attention weights are schematic.
    \textbf{b,c,d,} ISUP grading performance (quadratic-weighted $\kappa$) versus training-set size for UNI2-h, Virchow2 and CONCH, comparing ABMIL (grey), the CLAM-inspired pseudo-label control (labelled CLAM; blue) and CoworkMIL with agent-generated ROIs (purple). Lines show means over 20 matched training seeds on a fixed 30-slide test set; bands show standard errors calculated across seeds (Methods). Stars denote paired, two-sided exact Wilcoxon signed-rank tests against ABMIL: *$P<0.05$, **$P<0.01$. Exact $P$ values are listed in Supplementary Table~\ref{tab:mil_paired}.
    \textbf{e,} Difference in $\kappa$ between CoworkMIL and ABMIL across training-set sizes for each backbone.
    \textbf{f,} Comparison of the three methods at 20 training slides; bars show means over 20 seeds, and error bars use the same standard-error calculation as in \textbf{b,c,d}.
    \textbf{g,} Effect of auxiliary-loss weight $\beta$ on CoworkMIL performance (UNI2-h, 20 training slides), averaged over ten seeds. These are the first ten seeds of the main comparison; the dashed line denotes their ABMIL mean.
    \textbf{h,} Illustrative prostate biopsy with CoworkMIL and ABMIL attention maps and the 16 highest-attention patches from each model.}
    \label{cap:fig:coworkmil}
    \caption*{Artwork on page~\pageref{fig:coworkmil}.}
\end{figure}

\subsection*{PathCowork builds pathology AI benchmarks from recorded clinical practice}

Beyond training AI models, PathCowork also enables the construction of case-specific benchmarks from recorded clinical examinations, allowing us to evaluate whether AI models can understand the visual context, interpret evidence, and make decisions encountered during real-world pathology practice.
Specifically, we use PathCowork to generate case-specific questions from the pathologist's observations and viewing history, covering four demands: integrating evidence across slides, rejecting a question with an inapplicable premise, identifying the tissue location or viewing magnification, and choosing where to look next for a stated goal (\textbf{Figure~\ref{fig:benchmark}a}). We formally named this benchmark as ``CoworkExam''.
For example, a recorded recommendation to request C-kit staining supplies the answer to a question about which immunostain to order after reviewing several slides. Other questions require recognizing that a field does not contain the tissue assumed by the question or selecting a region for closer examination (\textbf{Supplementary Figure~\ref{fig:benchmark_examples}}). A separate text-based analysis examined how much of the recorded checking content was retained by generated questions, compared with questions selected from PathMMU and textbook-derived banks (\textbf{Supplementary Figure~\ref{fig:recorded_check_coverage}}; Supplementary Note~\ref{supp:outputs}).

We compared questions drawn from recorded cases with questions from PathMMU and a dermatopathology textbook (\textbf{Figure~\ref{fig:benchmark}b}; Supplementary Note~\ref{supp:outputs}).
On the same 148 CoworkExam questions, Opus 4.8 answered 39 correctly without images (26.4\%) and 121 with images (81.8\%), a difference of 55.4 percentage points. The displayed PathMMU scores were 59\% and 74\%, respectively (15 points).
The displayed GPT scores were 23\% and 55\% on CoworkExam (32 points), versus 62\% and 70\% on PathMMU (8 points).
Image-supported accuracy was higher for both question sources, with larger source-level differences on CoworkExam.

A separate comparison across Haiku, Sonnet and Opus showed a wider range of scores on CoworkExam questions: 41\%, 52\% and 77\%, respectively, versus 56\%, 73\% and 74\% on PathMMU and 32\%, 30\% and 33\% on textbook questions (\textbf{Figure~\ref{fig:benchmark}b}).
The Sonnet-to-Opus difference was 25 percentage points on CoworkExam and 1 point on PathMMU.
Recorded-review questions yielded a wider spread of scores across these model configurations.

Across the four categories, six frontier models and one pathology resident showed distinct capability profiles (\textbf{Figure~\ref{fig:benchmark}c}).
The resident scored 91\% on rejecting inapplicable premises, compared with 13--79\% for the models, and 84\% on environmental sensing, compared with 46.5--79\%.
By contrast, the highest model scores were close to the resident's on multi-slide reasoning (89\% versus 87\%) and goal-directed navigation (90\% versus 89\%).
The largest remaining gaps were in recognizing the visual context of an examination and identifying when a question was based on an incorrect premise, highlighting the need for more rigorous pathology AI benchmarks and better reasoning models that can learn from real-world clinical practice, as enabled by PathCowork.

\clearpage
\newgeometry{left=1.5cm,right=1.5cm,top=1.0cm,bottom=1.0cm}
\begin{figure}[p]
    \pdfbookmark[2]{Figure 5}{paper.figure.5}
    \centering
    \includegraphics[width=180mm,height=\dimexpr\textheight-32pt\relax,keepaspectratio]{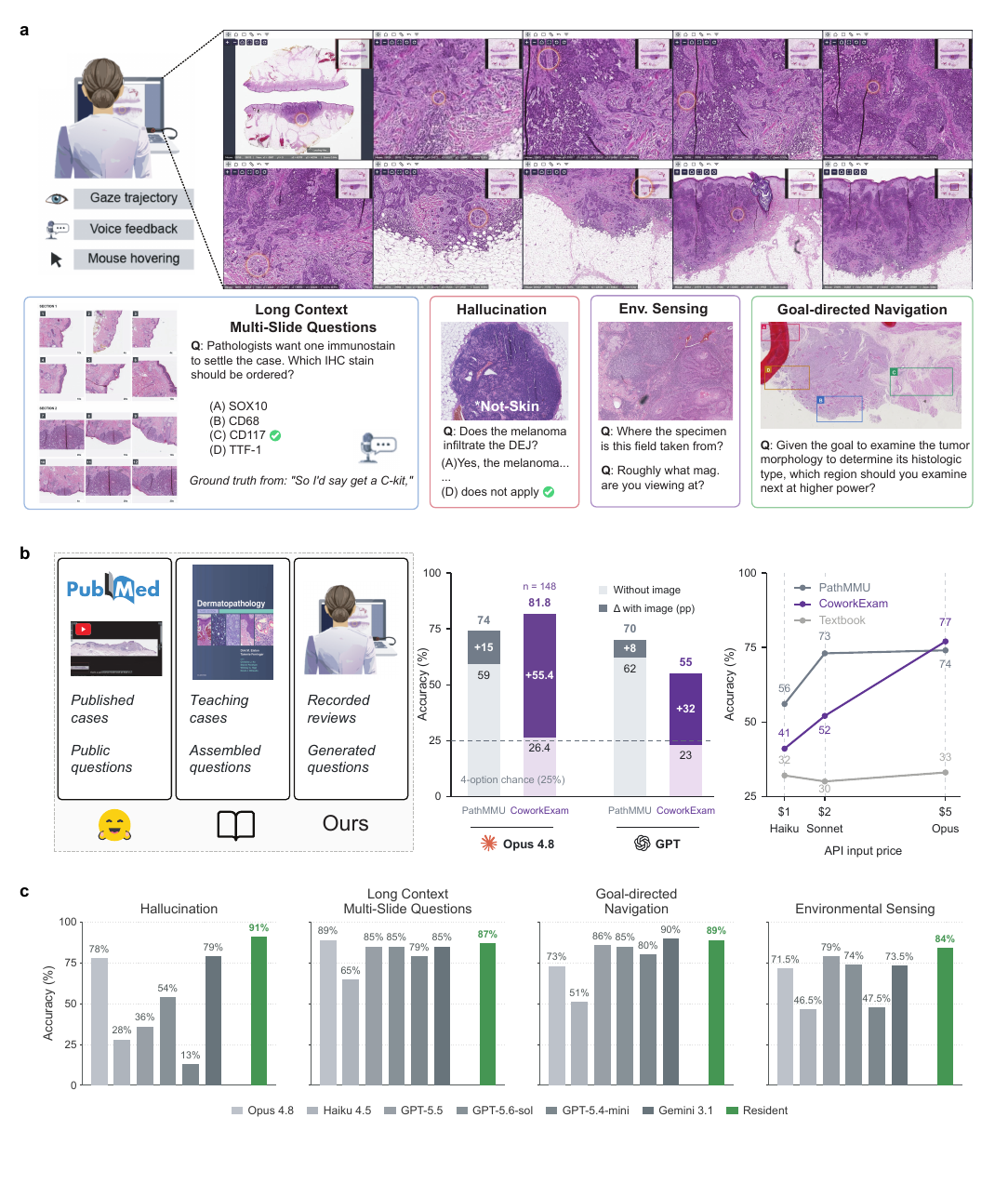}
    \caption{Legend on page~\pageref{cap:fig:benchmark}.}
    \label{fig:benchmark}
\end{figure}
\restoregeometry
\begin{figure}[H]
    \ContinuedFloat
    \caption{\textbf{Evaluating models using the context of recorded examinations.}
    \textbf{a,} Generation of case-specific questions from review histories, with examples of multi-slide reasoning, rejection of an inapplicable premise, environmental sensing and goal-directed navigation.
    \textbf{b,} Left, schematic comparison of question sources. Middle, displayed accuracy without images (lower bar segment) and the additional score with images (upper segment) for Claude and GPT on PathMMU and PathCowork (CoworkExam); the total bar height gives accuracy with images and the dashed line marks the 25\% reference for a four-option item with one accepted answer. Segment annotations give percentage-point differences. The Opus 4.8 comparison on CoworkExam uses the same 148 questions in both input conditions (39/148 without images; 121/148 with images). Right, accuracy on the separate 100-item sets per source across Haiku, Sonnet and Opus on PathMMU, PathCowork and textbook questions; the horizontal axis shows the input-price values used in the source comparison. The source schematic shows the provenance of the question materials.
    \textbf{c,} Accuracy of six frontier models and one pathology resident across four question categories: rejection of an inapplicable premise, multi-slide reasoning, goal-directed navigation and environmental sensing.}
    \label{cap:fig:benchmark}
    \caption*{Artwork on page~\pageref{fig:benchmark}.}
\end{figure}

 
\section*{Discussion}
\pdfbookmark[1]{Discussion}{paper.discussion}

Clinical AI has largely learned from static images and final diagnostic reports, leaving the interpretive process that connects them unrecorded. PathCowork addresses this gap by capturing how pathologists examine tissue, link spoken observations to specific image regions and viewing scales, and build diagnostic explanations. Treating clinical interpretation itself as the primary source of supervision marks a fundamental change in how medical experience becomes AI learning, moving beyond predicting endpoint labels to capturing how experts reason through visual evidence during diagnosis.

This work has several strengths. Most fundamentally, it transforms routine diagnostic experience into durable, reusable knowledge, preserving the reasoning that normally vanishes once a report is signed. We developed an evidence-localization grounder that outperformed frontier commercial models. Agents combining this grounder with recorded diagnostic procedures were more accurate, faster and substantially cheaper to run than Codex and Claude Code. Furthermore, the value of this captured expertise extends beyond agents: using agent-identified regions to guide slide-level model training when labelled slides are limited shows that recorded clinical reasoning can become transferable supervision across computational tasks. Finally, this shift from static answers to active reasoning changes how medical AI should be judged. Through CoworkExam, evaluation moves beyond factual recall or disease naming to test interpretation in context, including whether a model can recognize the viewing context, reject inapplicable premises, integrate evidence across slides and decide what to examine next.

This work has several limitations. First, the current evaluations measure the standalone diagnostic accuracy of PathCowork, establishing the performance of the agent alone. In the foreseeable future, however, PathCowork is intended to serve as a copilot for pathologists, and how the system affects clinicians during routine diagnosis remains unstudied. Prospective studies of real clinical use, although out of the scope in this work, will be needed to determine whether pathologists using PathCowork work faster, diagnose more accurately, gather more complete evidence and experience less subjective fatigue over the same duration of work, while also assessing their perceived usability and overall experience with the system. Longitudinal follow-up should also examine whether sustained use maintains or erodes clinicians' unaided diagnostic proficiency.

Second, the current recordings capture the complete traditional diagnostic workflow of a pathologist directly examining tissue images and interpreting findings. Increasingly, however, pathologists are adopting AI tools, such as automated cell counting or mitosis detection, to accelerate their work. Extending PathCowork from two-way interactions between pathologists and images to three-way interactions among pathologists, images and AI tools would capture how clinicians use and interpret AI outputs during diagnosis, enabling future agents to learn from these interactions.

Third, learning from recorded examinations is currently passive, so diagnostic concepts may receive uneven supervision depending on how often they arise during routine cases. A partly active learning approach could allow an agent to recognize concepts or subtle morphological distinctions it has not yet learned well and ask targeted questions, request demonstrations or seek corrections during an examination, focusing supervision where needed and adapting the system to an individual pathologist's specialized expertise.

Looking ahead, these capabilities point toward AI systems that could shadow pathologists during routine diagnosis, learning mostly by observation while asking targeted questions when unfamiliar or ambiguous findings arise. Paired with rigorous benchmarks such as CoworkExam to test whether an agent has learned to interpret visual evidence in context and act appropriately, this direction offers a foundation for transparent, personalized clinical copilots shaped directly by the experts they work alongside.
 
\section*{Methods}
\pdfbookmark[1]{Methods}{paper.methods}
\newfloat{methodfigure}{htbp}{lom}
\floatname{methodfigure}{Methods Figure}
\renewcommand{\themethodfigure}{M\arabic{methodfigure}}
\providecommand{\theHmethodfigure}{methods.\arabic{methodfigure}}
\newlength{\methodfigurewidth}
\newcommand{\methodfigureblock}[4][16cm]{%
\begin{methodfigure}[H]
\pdfbookmark[2]{Methods Figure M\number\numexpr\value{methodfigure}+1\relax}{bookmark.#4}
\centering
\setlength{\methodfigurewidth}{#1}%
\ifdim\methodfigurewidth>\linewidth\setlength{\methodfigurewidth}{\linewidth}\fi
\includegraphics[width=\methodfigurewidth]{#2}
\captionsetup{width=\linewidth,name=Methods Figure}
\caption{#3}\label{#4}
\end{methodfigure}}
\newcommand{\methodfig}[1]{\csname methodfigure:#1\endcsname}

\expandafter\newcommand\csname methodfigure:fig:ivt_idt\endcsname{%
\methodfigureblock[17cm]{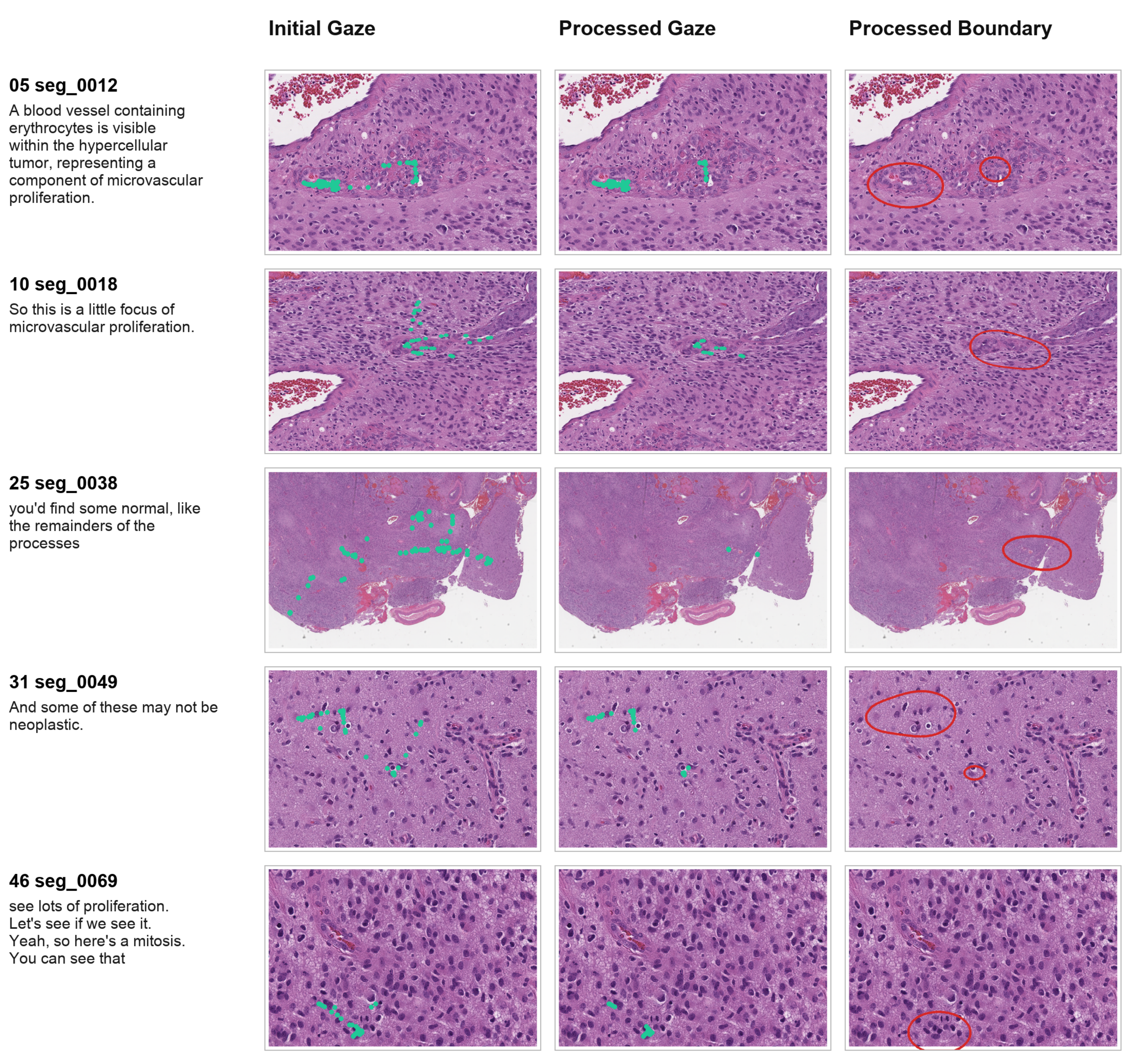}{\textbf{Gaze processing and extraction of attention regions.} Each row shows a recorded segment and its aligned narration. Columns show the initial gaze samples, samples retained after velocity- and dispersion-based fixation filtering, and the resulting attention boundaries over the viewed tissue. The side-by-side views allow the extracted regions to be inspected against the tissue and narrated findings, including microvascular proliferation and mitosis in glioblastoma.}{fig:ivt_idt}}

\expandafter\newcommand\csname methodfigure:fig:review\endcsname{%
\methodfigureblock[13cm]{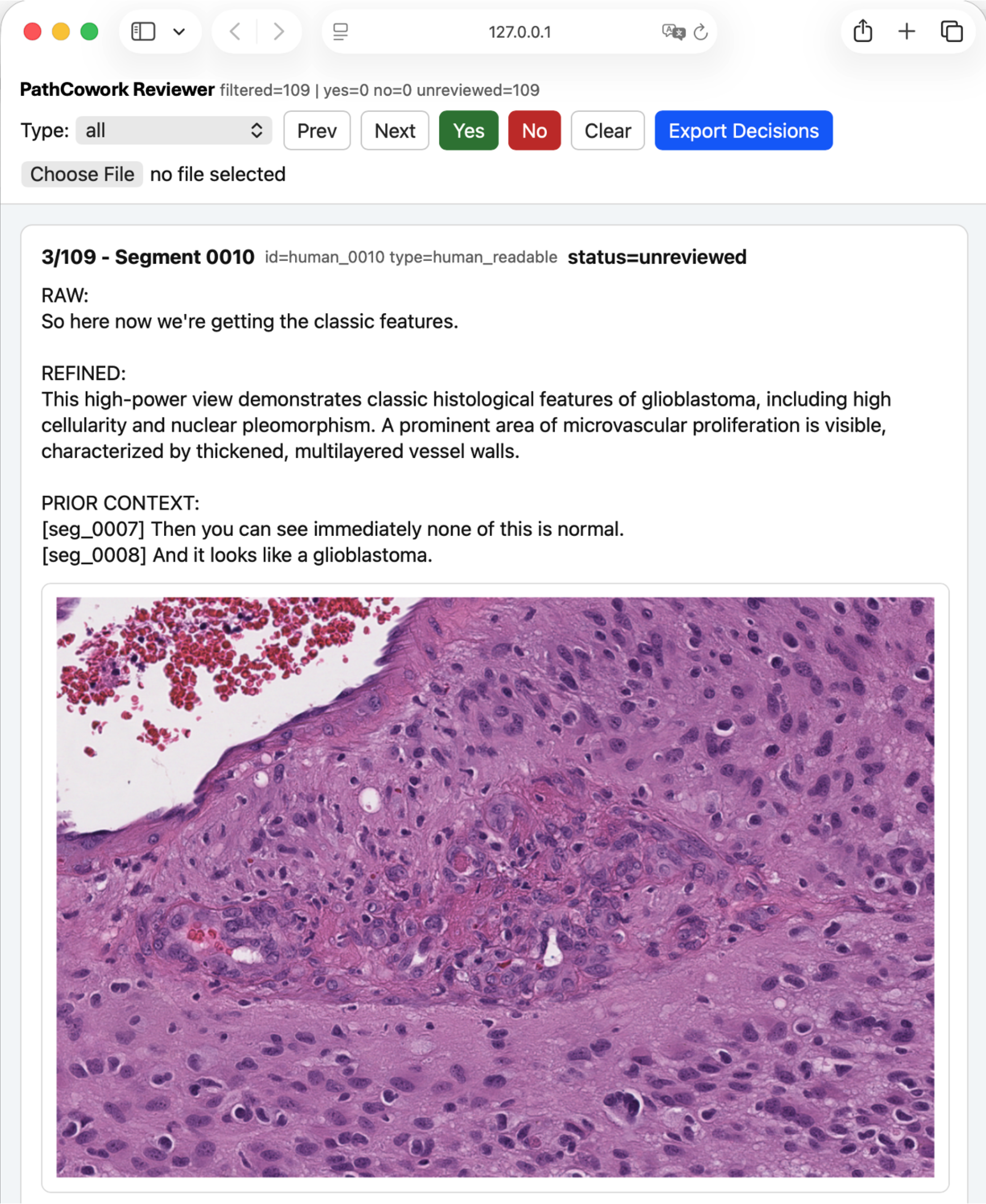}{\textbf{Interface for reviewing generated examples.} The reviewer displays the original utterance, the generated description, preceding narration and the corresponding image. Controls record approval or rejection and export the review decisions. The screenshot illustrates an unreviewed example.}{fig:review}}
 
\subsection*{Study design and recorded corpus}
The study was approved by the Institutional Review Board of the Perelman School of Medicine at the University of Pennsylvania (IRB \#857258), and all participants provided informed consent. Recording took place from February 2025 to September 2026. The recorded corpus comprised 48 sessions and 1,028\,min of review from 16 pathologists: six attending physicians, four fellows and six residents. These 16 participants included 12 co-authors of this study and four pathologists whose recordings were reused from our previous Pathology-CoT study~\cite{wang2026pathologycot}. Participants reviewed diagnostic or teaching material while explaining their observations as they would during a discussion at a microscope. Eye-tracker calibration preceded recording. Recordings with unusable audio were excluded from transcription-based processing.

A session denotes one continuous sitting at the WSI viewer; it can contain several slides, and a slide can be revisited. Sessions, slides and extracted viewing segments therefore represent different units of observation. The image collections span skin, prostate, breast, brain, lymphoid, gastrointestinal, bladder and cytology material (Supplementary Table~\ref{tab:datasets}). Public recording sources included The Cancer Genome Atlas (TCGA)~\cite{tcga2013pancancer}, DROID-skin~\cite{lindman2019droidskin,stadler2021database} and LNCO2~\cite{maras2020lnco2}. Recordings were processed after the session to produce learning material.

The recordings supplied grounding-training examples, grounding-evaluation examples and material for skill extraction. Grounding training and evaluation were separated at the patient level. Skills and complete agents were evaluated on unseen cases outside the recordings used for grounding training and skill construction. Benchmark construction drew on the full recorded corpus; benchmark evaluation therefore included external models and a pathology resident, without evaluating the PathCowork models.

\subsection*{Recording platform and synchronization}
PathCowork records slide review and narration on a common timeline. The platform extends the TissueLab whole-slide image (WSI) viewer~\cite{li2025co}, with a Next.js browser interface and an Electron desktop application. Viewport coordinates, magnification, mouse position, gaze and audio are recorded while a pathologist examines a slide and explains the findings.

Gaze coordinates enter the viewer through a WebSocket connection to a calibrated eye tracker. Mouse and viewport events are captured through viewer hooks and browser event listeners, and audio through the media-capture interface. Mouse and viewport state are polled at 10\,Hz, and the gaze buffer is read at 20\,Hz; the tracker sampling frequency is distinct from this readout rate. Events are timestamped against the workstation clock to align speech with the visual field. Alignment resolution depends on polling, buffering and transcription timing.

Sessions were recorded on Windows workstations with the available eye-tracker drivers. The software supports local and server-backed storage. Screen recordings allow the event streams to be checked against the slide displayed during review. Subsequent processing operates on the saved streams.

\subsection*{Processing viewing events and speech}
The event-processing pipeline groups pan, zoom and dwell records into viewing episodes. Low- and intermediate-power events capture tissue survey, whereas high-power events capture examination of individual cellular features. Survey views are binned to 5$\times$ or 10$\times$ equivalents. A stationary view lasting more than 1\,s, or continuous panning lasting more than 2\,s at these scales, defines an inspection event. High-power views yield a 1,024$\times$1,024-pixel crop centered on the viewport. Brief navigation events are retained in the trajectory.

Audio is transcribed with Whisper-large-v3-turbo~\cite{radford2023whisper} using pathology vocabulary and slide context as prompts. Segment timestamps link each utterance to the corresponding viewing episode. A Qwen3-4B text classifier~\cite{yang2025qwen3} receives the utterance together with two preceding and two following segments to distinguish diagnostic observations from interface discussion, off-topic speech and other non-diagnostic content. The raw transcript is retained alongside the normalized description for checking against the original speech.

\subsection*{Gaze filtering and spatial association}
Gaze samples are mapped from screen coordinates into WSI coordinates using the recorded viewport. A hybrid velocity- and dispersion-based procedure identifies fixations~\cite{salvucci2000fixations}. The velocity stage uses a threshold of 1,800\,pixels\,s$^{-1}$ in the mapped image coordinates. The dispersion stage uses a 220-pixel threshold and a minimum duration of 90\,ms, with dispersion calculated as the sum of horizontal and vertical coordinate ranges. Each retained fixation carries its start and end times, duration, centroid, sample count and dispersion.

For each speech-aligned region, retained samples are rasterized into a two-dimensional density map, smoothed and thresholded to obtain contiguous attention regions. These regions can be represented as boundaries or polygons in WSI coordinates (\textbf{Methods Figure~\ref{fig:ivt_idt}}). These boundaries represent the clinician's sampled visual attention.

\methodfig{fig:ivt_idt}

\subsection*{Cross-modal association and review}
The pipeline forms candidate regions from gaze fixations, mouse-position density and viewport changes concurrent with speech. An utterance can have fewer than three usable candidates, depending on the recorded behavior. Qwen3.5-397B-A17B~\cite{yang2025qwen3} receives the image, candidate overlays and aligned narration and selects the region that best matches the described finding. The selected geometry, source modality and narration form a localized training example.

Nearby viewing events at similar magnifications are merged to simplify the navigation trajectory. Returns to a previously examined region are retained as revisits. A review interface presents the crop, transcript, generated description and preceding context together. Reviewers can check their correspondence and export decisions for downstream filtering (\textbf{Methods Figure~\ref{fig:review}}).

\methodfig{fig:review}

Each processed example retains its source image, original narration, timestamps, WSI coordinates and viewing scale. Generated descriptions and contextual summaries accompany the recorded speech; fields for uncertainty and interpretation are inferred from that narration. Trajectories preserve the sequence of views and link repeated visits. These links make derived examples traceable to the original slide and utterance. Detailed fields are listed in Supplementary Table~\ref{tab:data_schema}.

\subsection*{Descriptive analysis of recorded behavior}
For the narration composition analysis, an LLM assigned each sentence one primary category using the preceding sentence as context. Among diagnostic-review recordings, 7,638 sentences were categorized; 2,879 assigned to ``other'' were excluded, leaving 4,759 pathology-related sentences. This categorization is separate from the diagnostic-content filter used during record processing. Vocabulary enrichment was calculated from segment-level presence or absence of each lexeme for each skin reader versus the pooled non-skin records. The log odds ratio used a 0.5 correction in all four contingency-table cells and was divided by its estimated standard error. Displayed terms were selected exploratorily for consistent directions across readers.

Concepts were identified by lexical matching and could overlap within a segment. Objective-equivalent viewport magnification was calculated as the slide's base objective multiplied by $\sqrt{(2047\times1440)/(w h)}$, where $w$ and $h$ are the viewport dimensions in slide pixels. Per-slide optics metadata supplied the base objective; unreliable rotated viewport boxes were excluded. For the spatial-term analysis, gaze points were pooled from whole viewing segments overlapping an utterance and its surrounding 5\,s. Windows containing fewer than three points were excluded. RMS spread was the square root of the mean squared distance from the window centroid, converted to millimeters using per-slide pixel spacing. Centered windows were pooled by term with equal weight per gaze sample, binned within $\pm6$\,mm and smoothed with a 300\,$\mu$m Gaussian. Contours enclose 20\%, 50\% and 80\% of the smoothed in-field mass. These analyses are descriptive: segments and overlapping gaze windows within a reader are not independent observations.

\subsection*{Construction of executable skills}
Skill construction begins with a target clinical task. Task names are expanded into related diagnostic terms and used to retrieve relevant recorded narration and case context. For the melanoma-versus-dysplastic-nevus task, these terms include atypical nevus, melanocytic neoplasm and melanocytic proliferation. The retrieved material is pooled across the participating dermatopathologists.

Narration is organized into an ordered procedure, such as finding the dermoepidermal junction, examining junctional growth, assessing the dermal component and searching for mitoses. Each procedural concept is then linked to supporting image regions and the magnifications at which those regions were examined. The available spatial representation can be a polygon, bounding box or polyline, depending on the finding. Repeated retrieval and grounding connect the procedural steps to examples from the recordings.

The resulting \texttt{skill.md} specifies the target anatomy, viewing scale, action and decision context for each step. The execution sequence implements the examination order and viewing scales extracted from the recordings. The skill guides an agent that can navigate a WSI and invoke a trained localizer; supervised fine-tuning on the associated behavioral examples teaches the localizer to identify the concepts required by the procedure.

\subsection*{Supervised concept grounding}
The main grounding comparison evaluates Kimi K2.6~\cite{moonshot2026kimik26} and Qwen3.6-27B~\cite{yang2025qwen3} before and after supervised fine-tuning on recorded examples. The adapted models are named PathCowork-1T and PathCowork-27B, respectively. Each example pairs an image and concept query with a structured spatial response. The detailed dermatopathology grounding evaluation in \textbf{Figure~\ref{fig:behavior}e,f} and the agent in \textbf{Figure~\ref{fig:skill}} use the PathCowork-1T localizer. Additional Qwen development examples retain their own model configuration (Supplementary Note~\ref{supp:training}); the atlas-derived glioma adaptation study below is a separate experiment. Model identifiers are reported by experiment in Supplementary Table~\ref{tab:model_settings}.

The two main adaptation runs used rank-32 LoRA~\cite{hu2022lora}, a peak learning rate of $10^{-4}$, batch size 16 and two training epochs. We used the checkpoint saved at the end of training. Separate development experiments examined model size, numerical precision and training duration to characterize the trade-offs between localization and deployment requirements. The 4B experiments examine spatial-format and reinforcement-learning choices at a smaller scale.

Training examples and reference geometries were derived from recorded viewing behavior and aligned narration, with gaze-derived candidate regions extracted from filtered fixations (\textbf{Methods Figure~\ref{fig:ivt_idt}}). Gemini 3.1 screened candidate references for obvious errors, followed by manual spot-checking by a pathologist. Erroneous examples were removed or corrected. Data augmentation and automatic generation of negative examples supplemented the behavior-derived examples. Test examples were constructed using the same procedure, with 50 image--concept pairs per evaluated concept and patient-level separation from the training data. The seven-concept dermatopathology evaluation therefore comprised 350 image--concept pairs (Supplementary Note~\ref{supp:training}).

In the main grounding evaluation, missed targets and unparseable spatial responses receive a localization score of zero and remain in the evaluation denominator. For the aggregate grounding evaluation in \textbf{Figure~\ref{fig:behavior}d}, spatial agreement is measured by Dice, $2|P\cap R|/(|P|+|R|)$, where $P$ and $R$ are predicted and reference masks. Polygons are rasterized as filled masks. Predicted and reference polylines are each expanded into a band of the same 50-pixel width before mask comparison. Dice scores are first averaged within each organ. The aggregate score is the mean of these organ-level averages, and error bars show their standard deviation. Detection recall is the number of correctly detected positive examples divided by the number of reference-positive examples; precision is the number of correctly detected positive examples divided by the number of predicted-positive examples.

Concept evaluations distinguish detection from spatial agreement with the reference region. We report detection recall and precision separately from grounding against regions and contours derived from pathologist behavior. Additional prostate analyses measure localization IoU and the proportion of outputs that can be parsed. We varied training-set size to examine overall and concept-specific detection and localization, and compared fixed-step with equal-epoch training budgets (\textbf{Supplementary Figures~\ref{fig:scaling_mech}, \ref{fig:perconcept} and~\ref{fig:scaling_compute}}; Supplementary Note~\ref{supp:scaling}). Comparisons with frontier models include aggregate prostate results and individual fields (\textbf{Supplementary Figures~\ref{fig:baseline}, \ref{fig:prostate_grounding_1} and~\ref{fig:prostate_grounding_2}}). Withholding cribriform examples assessed transfer to a concept without direct supervision (\textbf{Supplementary Figures~\ref{fig:holdout} and~\ref{fig:cribriform_examples}}). Separate configuration comparisons assessed backbone size, numerical precision and training duration (\textbf{Supplementary Figure~\ref{fig:model_choices}a,b,c}; Supplementary Note~\ref{supp:backbone}).

\subsection*{Spatial-format selection and glioma adaptation}
Before constructing supervised targets, we sampled the unadapted Qwen3.5-4B model under six prompts requesting different spatial formats. The comparison used 60 Ivy GAP~\cite{allen2015ivygap,puchalski2018ivygap} microvascular-proliferation tiles from eight held-out donors, including 40 positive tiles, at approximately 3\,$\mu$m per pixel and 2.3\,mm field width. These tiles were used for format development. Selection considered region production, presence discrimination and whether polygons could represent more than an axis-aligned rectangle. A region was classified as non-rectangular if it had more than four vertices or more than two distinct values along either coordinate axis. Localization was measured separately. The selected prompt contained one hypothetical JSON example; detailed format comparisons are reported in Supplementary Note~\ref{supp:recipe}.

Supervised inputs consisted of an image and a concept query, with a target JSON response containing polygons in normalized integer coordinates from 0 to 999, or an absent response for a negative example. Cross-entropy loss was computed only at answer-token positions; image and prompt positions had zero loss weight. We used rank-32 LoRA adaptation~\cite{hu2022lora}, Adam with learning rate $10^{-4}$, batch size eight and 768-pixel image inputs. Checkpoints were selected by validation negative log-likelihood using a donor-grouped split. The supervised initialization used for reinforcement learning was trained for two epochs.

Reinforcement learning was run from the base model and from the supervised checkpoint using the same pool of 2,553 records, including 426 negatives. A third condition used a separately supervised checkpoint and a pool augmented with 284 lower-grade glioma negatives. The corresponding supervised validation pools contained 77 records without augmentation and 117 with augmentation. The third condition changes both initialization and the reinforcement-learning prompt pool.

We used grouped response sampling following GRPO~\cite{shao2024deepseekmath}. At each of 100 reinforcement-learning steps, current weights were re-exported for sampling eight responses to each of 12 prompts at temperature 0.9. Each response's advantage was the unscaled difference between its reward and the mean reward of the eight responses to that prompt. Groups with identical rewards contributed no gradient. Each sampled batch received one importance-weighted policy-gradient update on response tokens, using rank-32 LoRA and learning rate $10^{-5}$. The objective was an unclipped, unregularized group-relative REINFORCE estimator~\cite{williams1992reinforce}, with the group mean as its baseline and spatial agreement as its reward.

For positive tiles, the reward was $0.05 I_{\mathrm{parse}}+0.10 I_{\mathrm{shape}}+0.85 F_1$, where the indicators denote a parseable response and at least one non-rectangular polygon. The spatial score $F_1=2PC/(P+C)$ combines purity $P$, the fraction of predicted area overlapping the atlas annotation, and coverage $C$, the fraction of annotated area recovered; it is zero when there is no overlap. For negative tiles, the reward was $0.05 I_{\mathrm{parse}}+0.85 I_{\mathrm{absent}}$. Masks were rasterized at $256\times256$ pixels. Positive and negative tiles therefore had reward maxima of 1.00 and 0.90, respectively. Validation localization and presence discrimination were monitored alongside sampled training reward; their trajectories and the format-development results are reported in Supplementary Note~\ref{supp:recipe}.

\subsection*{Scope of the system capability comparison}
\textbf{Figure~\ref{fig:main}m} compares the available implementations of PathCowork, Pathology-CoT~\cite{wang2026pathologycot}, SlideChat~\cite{chen2025slidechat}, PathChat~\cite{lu2024pathchat,modella2026pathchat2}, representative pathology encoders (UNI and CONCH)~\cite{chen2024uni,lu2024conch}, and the Claude Code and Codex configurations used in this study. PathChat and PathChat 2 are grouped as one family; the SlideChat entry refers to its released CVPR 2025 implementation.

Trainability denotes access to adapting the diagnostic model with additional training data; personal adaptation denotes a mechanism for learning from a clinician's recorded practice. Evidence localization denotes explicit spatial outputs such as regions, coordinates or annotations. Visible reasoning denotes inspectable examination steps, actions and associated evidence. Private deployment requires patient-facing inference to remain within the user's infrastructure. Partial support denotes a capability available in only part of the workflow or with limited output. The entries describe implementation capabilities. The dermatopathology application used recordings pooled across participants.

\subsection*{Agent-system evaluation}
The organ-level comparison evaluates PathCowork, Claude Code and Codex on five slide-review tasks: melanoma versus dysplastic nevus in skin biopsies (20 cases, 69 WSIs), prostate cancer detection and ISUP grading~\cite{epstein2016isup} (20 slides), Nottingham grading in breast resections (39 matched cases), glioblastoma versus lower-grade glioma in brain resections (30 cases, 15 per class), and metastasis detection in colon-resection lymph nodes (77 slides; ten positive and 67 negative). Evaluation images came from HISTAI~\cite{nechaev2025histai} (skin and breast), PANDA~\cite{bulten2022panda} (prostate), TCGA~\cite{tcga2013pancancer} (brain) and LNCO2~\cite{maras2020lnco2} (colorectal lymph nodes); evaluation units are listed in Supplementary Table~\ref{tab:agent_cohorts}. Agents inspect WSIs through slide-navigation tools and return task-specific diagnostic outputs. The PathCowork agent additionally uses a skill assembled from the recordings and a trained localizer.

PathCowork was run once per evaluation unit, defined as a case or WSI according to the task. Its scripted harness executes a finite set of spatial inspection steps at specified viewing scales, calling the grounder to identify the structures required by the skill. For example, junctional examination traverses the identified dermoepidermal junction at the prescribed scale. Execution terminates when the prescribed inspection steps have been completed; no additional agent-turn cap or task-level timeout was imposed on this workflow. An LLM summarizes the collected findings into the final diagnostic output. This completion rule differs from the iterative navigation used by the general-purpose agents.

The brain comparison used the same 30 cases for all three agents, with 15 cases per class. The two missing Claude Code reports were counted as incorrect in the 30-case accuracy and the corresponding class-recall denominator; PathCowork and Codex returned all 30 reports. Runs that did not complete within the time limit were counted as incorrect for task success. Elapsed time and API costs include failed runs through termination; a timed-out run contributes its full time limit and the API cost incurred during that run.

For the 20-slide prostate comparison, a Claude Code timeout is counted as a cancer-detection failure (17/20 detected). Its ISUP $\kappa$ and MAE are calculated from the 19 valid grade predictions, without imputing a grade for the timeout.

Colon node-level references were taken from the LNCO2 ground truth. Node detection recall measures the proportion of reference lymph nodes located by the agent. End-to-end per-node metastasis recall uses all ground-truth tumour nodes as its denominator and counts each node called positive as a true positive; missed tumour nodes count as false negatives. Per-node specificity measures the proportion of reference non-tumour nodes called negative. Slide-level detection is evaluated separately.

The overview uses one primary endpoint per organ: diagnostic accuracy for skin and brain, cancer detection for prostate, grade accuracy for breast, and node detection recall for colon lymph nodes. Secondary endpoints include melanoma recall and specificity, ISUP quadratic-weighted $\kappa$ and mean absolute error, breast component-score accuracy and grade-specific recall, class-specific brain recall, and colon per-slide recall and specificity. Detailed results, model configurations, task-specific denominators and reporting completeness are given in Supplementary Note~\ref{supp:agents} and \textbf{Supplementary Figure~\ref{fig:organ_supp}}. For breast, all three agents were evaluated on the same 39 cases (10 grade 1, 14 grade 2 and 15 grade 3). Overall-grade and component-score accuracies use 39 as the denominator. Grade-specific recall uses the number of reference cases of that grade. Costs are reported per slide at the recorded API prices; times are wall-clock values, with median time used for colon.

Cross-task summaries give each of the five tasks equal weight. The average performance gain is the arithmetic mean of the five task-specific percentage-point differences between PathCowork and each comparator. For cost and time, we first average the five reported task-level per-slide values for each system, then divide the comparator's average by the PathCowork average to obtain the cost-reduction factor or speedup. The time summary combines the reported mean times for four tasks with the median time for colon. Calculations use the source values before rounding for display.

\subsection*{ROI-supervised multiple-instance learning}
CoworkMIL transfers regions proposed by PathCowork on MIL training slides into auxiliary spatial supervision for a slide classifier. These regions are outputs of the concept grounder for prostate architectural patterns, including well-formed glands, fused glands, cribriform architecture, solid sheets and cords. Concept-specific regions are combined into binary spatial targets. The image encoder remains frozen while the pooling model learns from slide grades and binary region-of-interest (ROI) targets. At inference, the classifier uses image features alone, without running PathCowork or supplying ROIs.

From a 100-slide PANDA subset~\cite{bulten2022panda}, we constructed a 65-slide training pool and a fixed 30-slide test set, with five test slides in each of six categories (benign and ISUP grade groups 1--5). Training and test slide identifiers were disjoint. Cancer slides without available ROI annotations were preferentially allocated to the test set; cancer training slides required ROIs, whereas benign slides used empty masks. Subsets of 10, 20, 30, 40 or 50 training slides were sampled by cycling through shuffled grade-specific lists. Each comparison used 20 matched runs, sharing the training subset across methods within a run. Subset sampling used seeds 1000--1019, and model initialization and training order used seeds 0--19.

Non-overlapping 256-pixel level-0 tiles were retained at $\geq25\%$ tissue coverage, estimated by saturation-channel Otsu thresholding on slide thumbnails. Tiles were resized to 224 pixels with ImageNet normalization for UNI2-h and Virchow2, or to 448 pixels by bicubic interpolation with CLIP normalization for CONCH. Cached features comprised 1,536-dimensional UNI2-h~\cite{mahmoodlab2025uni2} class tokens; 2,560-dimensional Virchow2~\cite{zimmermann2024virchow2} embeddings concatenating the class token and mean patch token, excluding register tokens; and 512-dimensional pooled CONCH~\cite{lu2024conch} image features without contrastive projection or unit-length normalization. Binary tile targets indicated whether each tile center lay within the union of ROI polygons; polygon grade labels were ignored. Outside-ROI targets denote tissue unselected by the grounder.

A linear projection, ReLU and dropout (probability 0.25) mapped each tile embedding to 256 dimensions. A scalar linear suspicion head supplied both the tile-level prediction and the attention logit $s_i$. Softmax across tiles produced pooling weights, and a linear classifier mapped the weighted embedding sum to six slide categories. For slide label $y$, predicted class probability $p_y$ and binary targets $r_i$ for $m$ tiles, the effective objective was
\begin{equation}
\mathcal{L}=-\log p_y-\frac{\beta}{m}\sum_{i=1}^{m}\left[r_i\log\sigma(s_i)+(1-r_i)\log\bigl(1-\sigma(s_i)\bigr)\right],
\end{equation}
where $\sigma$ is the sigmoid function. The auxiliary binary cross-entropy averages all tiles within a slide, with $\beta=1$ throughout each main-comparison run. Slide-level cross-entropy was unweighted. AdamW used a learning rate and weight decay of $10^{-4}$. Models were trained for 40 epochs with shuffled slide order and one slide bag per update. Evaluation used the final-epoch checkpoint with dropout disabled and argmax prediction.

Comparators used the same frozen features, projection, slide head, optimization settings and training subsets. ABMIL~\cite{ilse2018attention} used gated attention pooling and slide-level cross-entropy alone. A CLAM-inspired pseudo-label control~\cite{lu2021clam} added binary supervision to the $k$ highest- and $k$ lowest-attention tiles, where $k=\max(1,\min(8,\lfloor m/2\rfloor))$. Its auxiliary loss was $\beta/2$ times the sum of the positive- and negative-group mean binary cross-entropies. This reimplementation used one shared binary instance head. CoworkMIL used the supervised tile logit for pooling as well as region prediction. The exploratory $\beta$ sensitivity analysis used the first ten matched runs at 20 training slides with UNI2-h, evaluating weights 0.1, 0.25, 0.5, 1 and 2 on the same test set. Exact paired comparisons against ABMIL are listed in Supplementary Table~\ref{tab:mil_paired}.

We separately evaluated EAGLE~\cite{neidlinger2026eagle}, a frozen attention-guided tile-selection pipeline, on the same 30 test slides and the same 20 training subsets of 20 slides. EAGLE used frozen CTransPath features~\cite{wang2022ctranspath} and CHIEF attention~\cite{wang2024chief} to select up to 25 tiles, then averaged their 1,280-dimensional Virchow2 class-token embeddings. All feature extraction within a run used either 0.5 or 2.0\,$\mu$m per pixel. Only a slide-level logistic-regression classifier was fitted to downstream labels, using raw embeddings, balanced class weights, L2 regularization ($C=1$), the \texttt{lbfgs} solver and a maximum of 10,000 iterations, following the released few-shot probe configuration. Results are summarized in Supplementary Table~\ref{tab:eagle_matched}.

\subsection*{Benchmark construction and evaluation}
Question generation proceeds from a case summary to individual viewing segments. Gemini-3-Flash first receives a 3$\times$3 montage, the session transcript and available case context to summarize the findings. It then receives individual segment images, attention overlays, aligned narration and the case summary to produce question--answer candidates. Questions and their accepted answers were manually reviewed against the source images, aligned narration and recorded viewing context. Items are scored by whether the selected option belongs to the accepted-answer set. Some items have multiple accepted options or a different number of options. The uniform random single-choice reference is therefore $|A_i|/|O_i|$ for item $i$, where $A_i$ and $O_i$ denote its accepted answers and available options; 25\% applies only to four-option items with one accepted answer.

The four categories address evidence integration across fields or slides, rejection of an inapplicable premise, recognition of the tissue or viewing context, and selection of the next region for a diagnostic goal. Narration and viewer state provide the evidence used to construct the intended answers. Models were evaluated against a shared question bank and accepted-answer keys. The retained bank contains 371 questions: 128 on inapplicable premises, 62 on multi-slide reasoning, 59 on goal-directed navigation and 122 on environmental sensing (63 viewing-scale and 59 tissue-location questions; Supplementary Table~\ref{tab:benchmark_composition}). Source-item examples are shown in \textbf{Supplementary Figure~\ref{fig:benchmark_examples}}.

The question-source comparison includes an image-input comparison for Claude and GPT on PathMMU~\cite{sun2024pathmmu} and PathCowork, and a separate comparison of Haiku 4.5, Sonnet 5 and Opus 4.8 across PathMMU, textbook-derived and PathCowork skin questions. Question-only inputs contain the question and options; image-supported inputs additionally contain the relevant image or montage. The textbook condition used locally assembled diagnosis questions from \emph{Dermatopathology: Diagnosis by First Impression}, fourth edition~\cite{ko2022dermatopathology}. The three-model source comparison uses a separate 100-item skin set per source. The four-category comparison includes Opus 4.8, Haiku 4.5, GPT-5.5, GPT-5.6-sol, GPT-5.4-mini and Gemini 3.1, together with one pathology resident. Accuracy is the proportion of evaluated questions answered correctly within each category. The archived multi-model scoring routine counts an attempted question without a parseable answer as incorrect, while excluding questions with no recorded attempt; evaluable denominators can therefore differ across models.

\subsection*{Statistical analysis}
Behavioral analyses summarize sessions, utterances and viewing segments at the level indicated in each panel. MIL performance is assessed with quadratic-weighted Cohen's $\kappa$ on six diagnostic categories. Learning-curve bands and fixed-size bars use the standard deviation across 20 seeds, calculated with divisor 20, divided by $\sqrt{20}$. The auxiliary-weight sweep instead summarizes ten seeds. Stars in \textbf{Figure~\ref{fig:coworkmil}b,c,d} compare CoworkMIL with ABMIL using paired, two-sided Wilcoxon signed-rank tests on matched seeds, without adjustment for multiple comparisons. All 15 comparisons, including their effect directions and exact $P$ values, are listed in Supplementary Table~\ref{tab:mil_paired}. Tests use the archived seed-level $\kappa$ values, stored to four decimal places, and quantify variation across training runs on the fixed test set.

Agent scores and cross-task summaries are descriptive, with each task contributing its specified clinical endpoint. The Opus 4.8 image-input comparison on PathCowork uses 148 matched questions with archived answers in both input conditions. These questions were selected from the recorded-review bank by retaining items with answers in all four archived Opus/Haiku image-supported and question-only conditions. The other displayed image-input comparisons retain their source-level scores. Question-level scores can include multiple items from the same recorded case.

\noindent\textbf{Author contributions.}
S. Wang and Z.H. conceived and designed the study. S. Wang, J.Y., S.L. and J.X. developed the software. M.V.R., S.K., E.B., M.N., W.D., C.M., T.C., A.W., Z.Y., D.A., S. Walle and X.X. contributed recorded slide reviews, including eye-tracking data and verbal descriptions. E.B. also assisted with early software debugging and refinement of the data collection protocol. O.P., E.B.L., Z.A., A.S., R.T., N.D., D.E. and G.J.N. provided advice on pathology domain knowledge. S. Wang and Z.H. wrote the manuscript. Z.H. supervised the study.

\noindent\textbf{Acknowledgements.}
This study was approved by the institutional review board (IRB \#857258) of the Perelman School of Medicine at the University of Pennsylvania.
We thank Jonathan Alex Miles, Annie Jin, Ikuko Hirai and Meaghan Dougher for contributing slide-review recordings to our previous Pathology-CoT study~\cite{wang2026pathologycot}, which were reused here.
We acknowledge Karin Lindman, Jer{\'o}nimo F. Rose, Martin Lindvall and Caroline Bivik Stadler for the DROID-skin dataset~\cite{lindman2019droidskin,stadler2021database}, and Gordan Maras, Martin Lindvall and Claes Lundstr{\"o}m for the LNCO2 dataset~\cite{maras2020lnco2}, provided through the AIDA Data Hub. The DROID-skin and LNCO2 datasets are copyright 2019 and 2020, respectively, Link{\"o}ping University, Claes Lundstr{\"o}m. We thank the Region G{\"a}vleborg Clinical Pathology and Cytology department for producing the LNCO2 collection.
We acknowledge the Ivy Glioblastoma Atlas Project (Ivy GAP)~\cite{allen2015ivygap,puchalski2018ivygap}, developed through a partnership of the Allen Institute for Brain Science, the Ben and Catherine Ivy Foundation, and the Ben and Catherine Ivy Center for Advanced Brain Tumor Treatment. The atlas was supported by the Ben and Catherine Ivy Foundation; our analyses and conclusions do not necessarily reflect the Foundation's views. Ivy GAP data are copyright 2015 Allen Institute for Brain Science and are available at \url{https://glioblastoma.alleninstitute.org/}.
The results published here are in part based upon data generated by the TCGA Research Network: \url{https://www.cancer.gov/tcga}. We thank the specimen donors and contributing research groups.

\noindent\textbf{Data availability.}
Example slide-review records, including segment metadata and images, are available in the \href{https://github.com/zhihuanglab/PathCowork/tree/main/history}{PathCowork history directory}.
The full training corpus and its complete concept inventory are not included in the model release.

\noindent\textbf{Code availability.}
PathCowork processing and review code is available on \href{https://github.com/zhihuanglab/PathCowork}{GitHub}. Links to model weights are also provided at \url{https://github.com/zhihuanglab/PathCowork}.

\noindent\textbf{Competing interests.}
The authors declare no competing interests.
\clearpage
\section*{Video legends}
\pdfbookmark[1]{Video legends}{paper.videos}

\phantomsection\label{video:agent}
\noindent\href{https://github.com/zhihuanglab/PathCowork/blob/main/videos/01_Main_Video_1_Agent.mp4}{\textbf{Main Video 1 | Interactive whole-slide review with PathCowork.}}

Screen recording of PathCowork examining a skin whole-slide image. After the user selects a slide and starts review, the agent surveys the tissue, localizes junctional features, inspects the dermis and candidate mitotic figures, and records findings with localized visual evidence before returning control to the user. The full 127.3-second recording is shown at its original speed, with cleaned audio of the original agent narration, English subtitles, and opening and closing title cards.

\medskip
\phantomsection\label{video:acquisition}
\noindent\href{https://github.com/zhihuanglab/PathCowork/blob/main/videos/02_Video_2_Behavior_Acquisition.mp4}{\textbf{Main Video 2 | Capturing pathologist behavior during skin-biopsy review.}}

Real-time screen recording of a pathologist examining a skin whole-slide image in TissueLab during a PathCowork data-acquisition session. The yellow ring displays gaze position as the reader pans and changes magnification. The original recorded discussion describes peripheral palisading, myxoid stroma and an infiltrative tumor edge, illustrating how spoken observations accompany the tissue being examined. The recording shows human review during data collection. English subtitles accompany the original audio. The full 48.1-s review is retained at its original speed; opening and closing title cards are added, and the audio level is increased for audibility.
 \newpage
\phantomsection
\bibliography{ref.bib}

\clearpage
\providecommand{\suppbookmarklevel}{1}
\providecommand{\suppchildbookmarklevel}{2}
\newgeometry{left=1.5cm,right=1.5cm,top=1.2cm,bottom=1.6cm}
\section*{Supplementary Information}
\pdfbookmark[\suppbookmarklevel]{Supplementary Information}{supp.information}
\setcounter{section}{0}
\setcounter{figure}{0}
\setcounter{table}{0}
\renewcommand{\thefigure}{S\arabic{figure}}
\renewcommand{\thetable}{S\arabic{table}}
\renewcommand{\theHfigure}{supp.\arabic{figure}}
\renewcommand{\theHtable}{supp.\arabic{table}}
\captionsetup[figure]{name=Supplementary Figure}
\captionsetup[table]{name=Supplementary Table}
\newcommand{\suppnote}[2]{\refstepcounter{section}\section*{Supplementary Note \thesection\enspace|\enspace#1}\pdfbookmark[\suppchildbookmarklevel]{Supplementary Note \thesection: #1}{bookmark.#2}\label{#2}}
\newcolumntype{L}[1]{>{\raggedright\arraybackslash}p{#1}}
\newlength{\suppfigurewidth}
\newcommand{\suppfigure}[4][16cm]{%
\begin{figure}[H]
\pdfbookmark[\suppchildbookmarklevel]{Supplementary Figure S\number\numexpr\value{figure}+1\relax}{bookmark.#4}
\centering
\setlength{\suppfigurewidth}{#1}%
\ifdim\suppfigurewidth>\linewidth\setlength{\suppfigurewidth}{\linewidth}\fi
\includegraphics[width=\suppfigurewidth]{#2}
\captionsetup{width=\linewidth}
\caption{#3}
\label{#4}
\end{figure}}
\newcommand{\suppfig}[1]{\csname suppfigure:#1\endcsname}

\expandafter\newcommand\csname suppfigure:fig:scaling_mech\endcsname{%
\suppfigure[15cm]{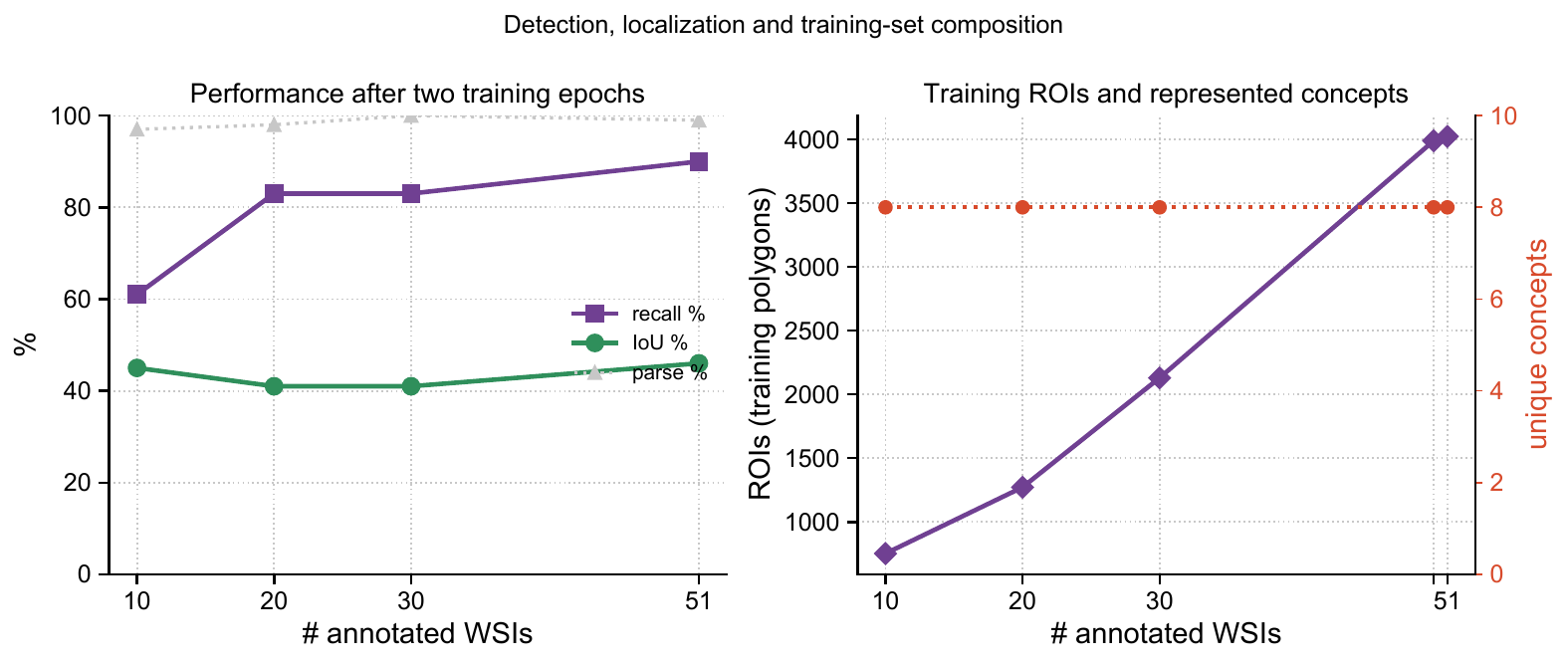}{\textbf{Data scaling of prostate concept grounding.} Left, detection recall, localization intersection over union (IoU) and output parse rate as the number of annotated training WSIs increases, with two training epochs at each size. Right, numbers of training polygons and represented concepts. The smallest set includes all eight concepts; the number of polygons increases from 752 to 4,024.}{fig:scaling_mech}}

\expandafter\newcommand\csname suppfigure:fig:perconcept\endcsname{%
\suppfigure[16.5cm]{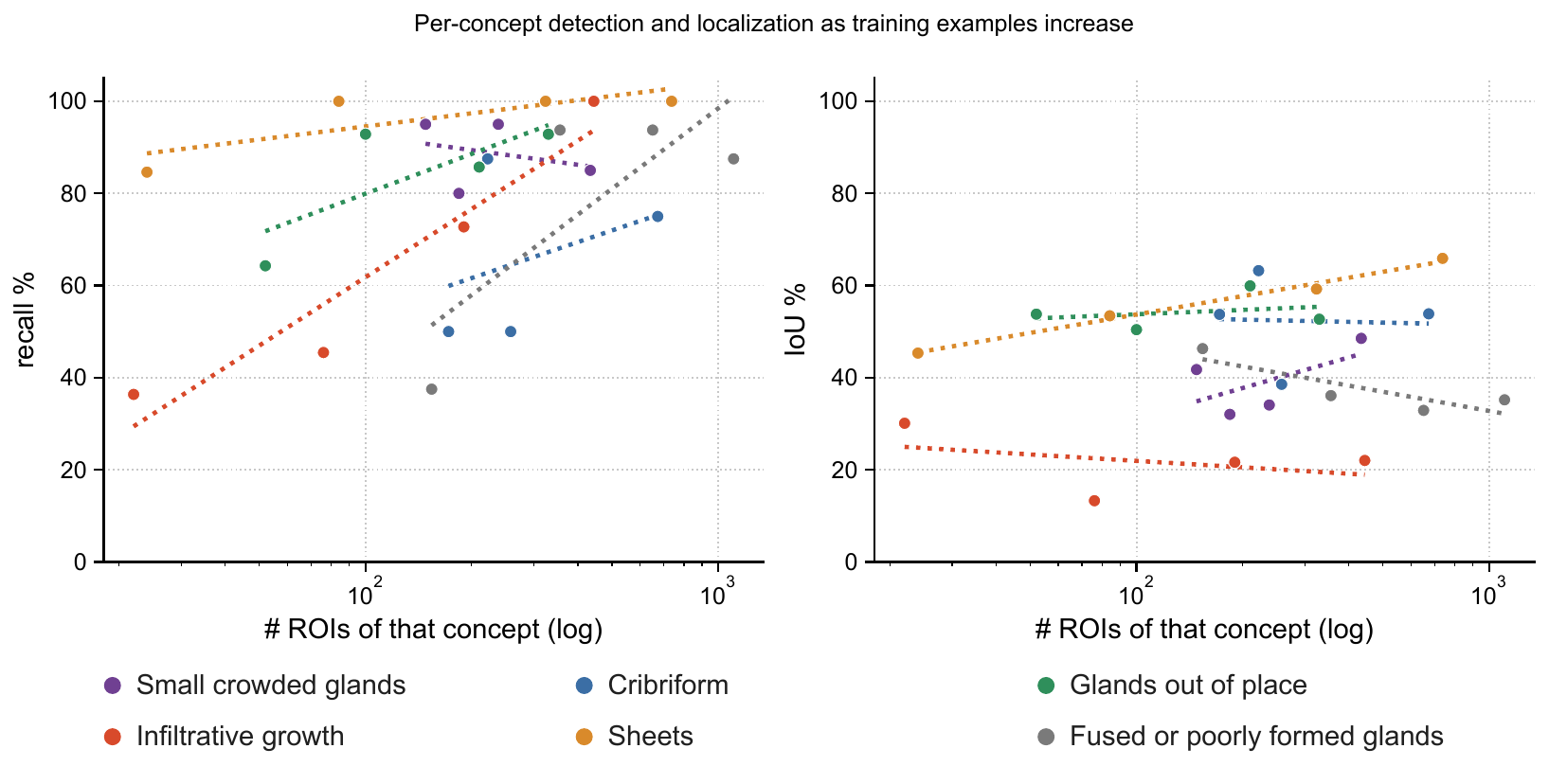}{\textbf{Detection and localization by concept and training-set size.} Recall (left) and localization IoU (right) are plotted against the number of training regions for six prostate architectural patterns: small crowded glands, cribriform architecture, glands out of place, infiltrative growth, sheets, and fused or poorly formed glands. Colors identify concepts; dotted lines show the fitted trends against log-transformed region counts.}{fig:perconcept}}

\expandafter\newcommand\csname suppfigure:fig:scaling_compute\endcsname{%
\suppfigure[15cm]{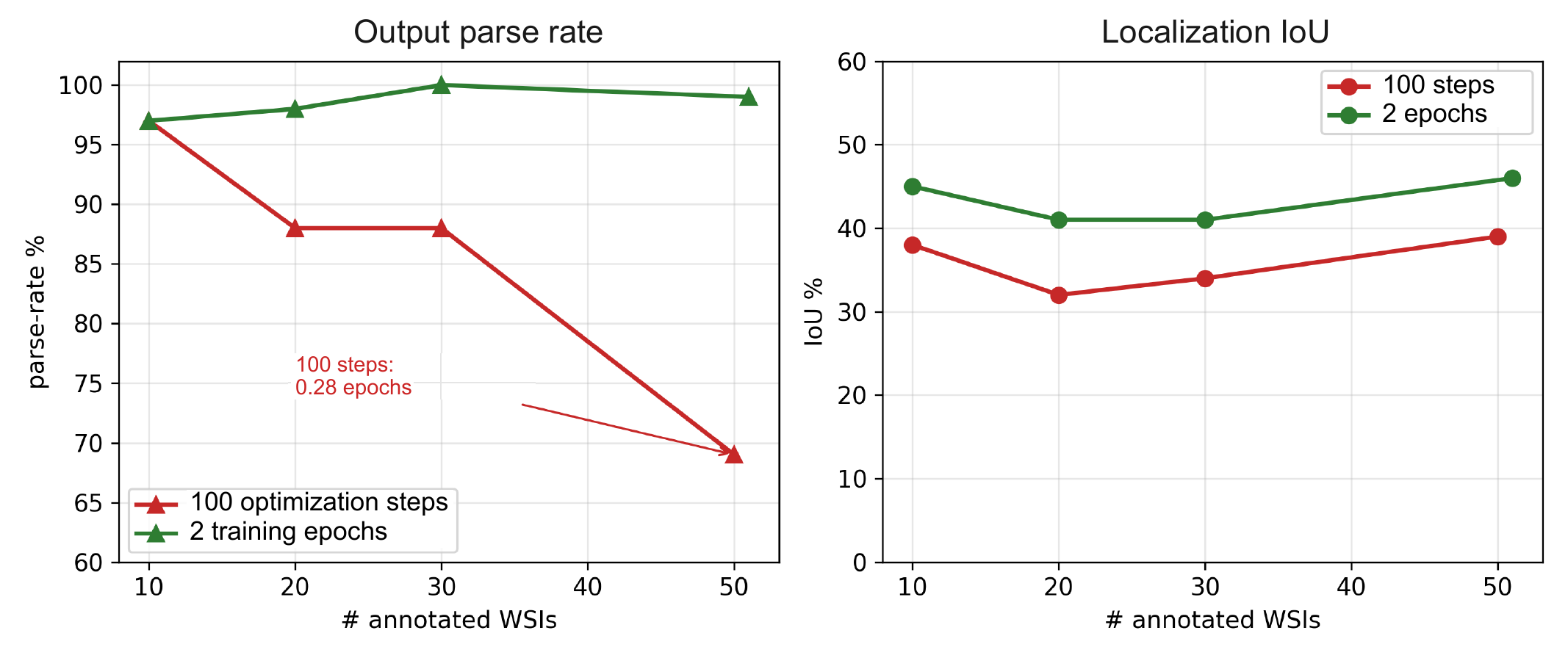}{\textbf{Fixed-step and equal-epoch training comparisons.} Output parse rate (left) and localization IoU (right) for a fixed budget of 100 optimization steps (red) and two epochs at each dataset size (green). The largest fixed-step run receives approximately 0.28 epochs.}{fig:scaling_compute}}

\expandafter\newcommand\csname suppfigure:fig:baseline\endcsname{%
\suppfigure[16cm]{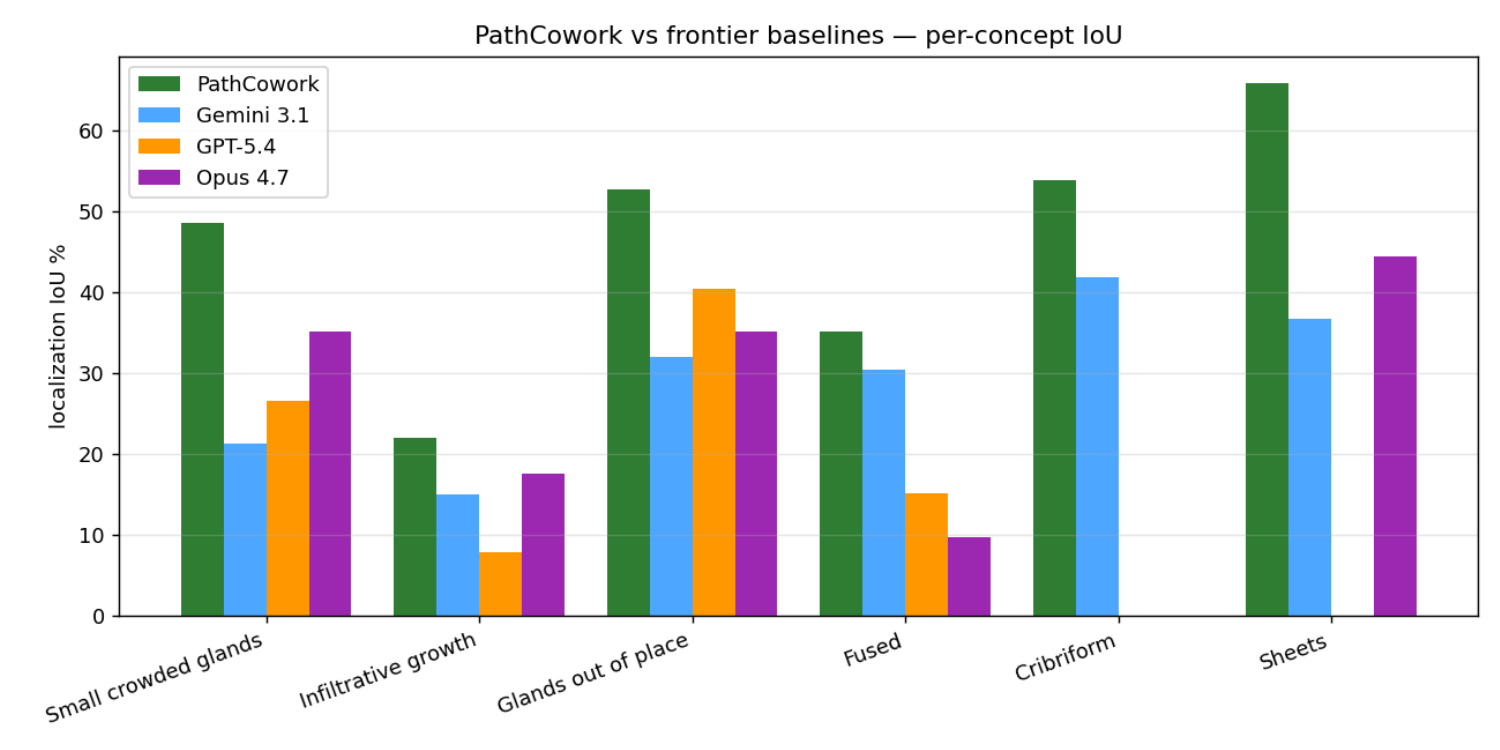}{\textbf{Localization of six prostate architectural concepts.} IoU for the adapted PathCowork localizer and Gemini 3.1, GPT-5.4 and Opus 4.7. Concepts are small crowded glands, infiltrative growth, glands out of place, fused or poorly formed glands, cribriform architecture and sheets.}{fig:baseline}}

\expandafter\newcommand\csname suppfigure:fig:holdout\endcsname{%
\suppfigure[12.5cm]{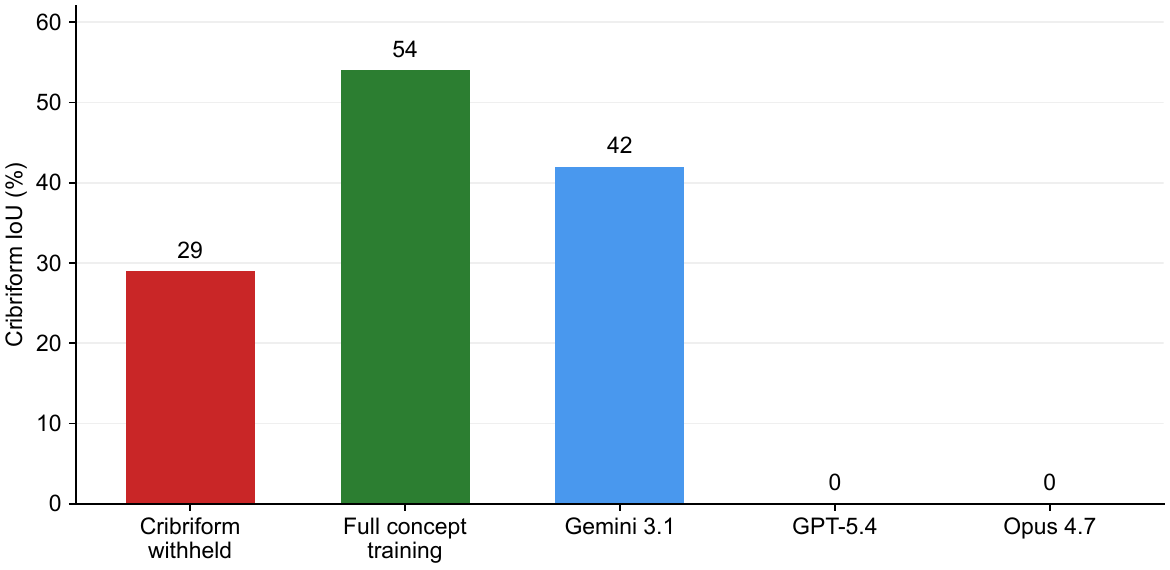}{\textbf{Localization after withholding cribriform supervision.} Cribriform IoU for a localizer trained without cribriform examples, the localizer trained with the full concept set, and three frontier baselines. Labels above the bars give IoU percentages.}{fig:holdout}}

\expandafter\newcommand\csname suppfigure:fig:model_choices\endcsname{%
\suppfigure[16.5cm]{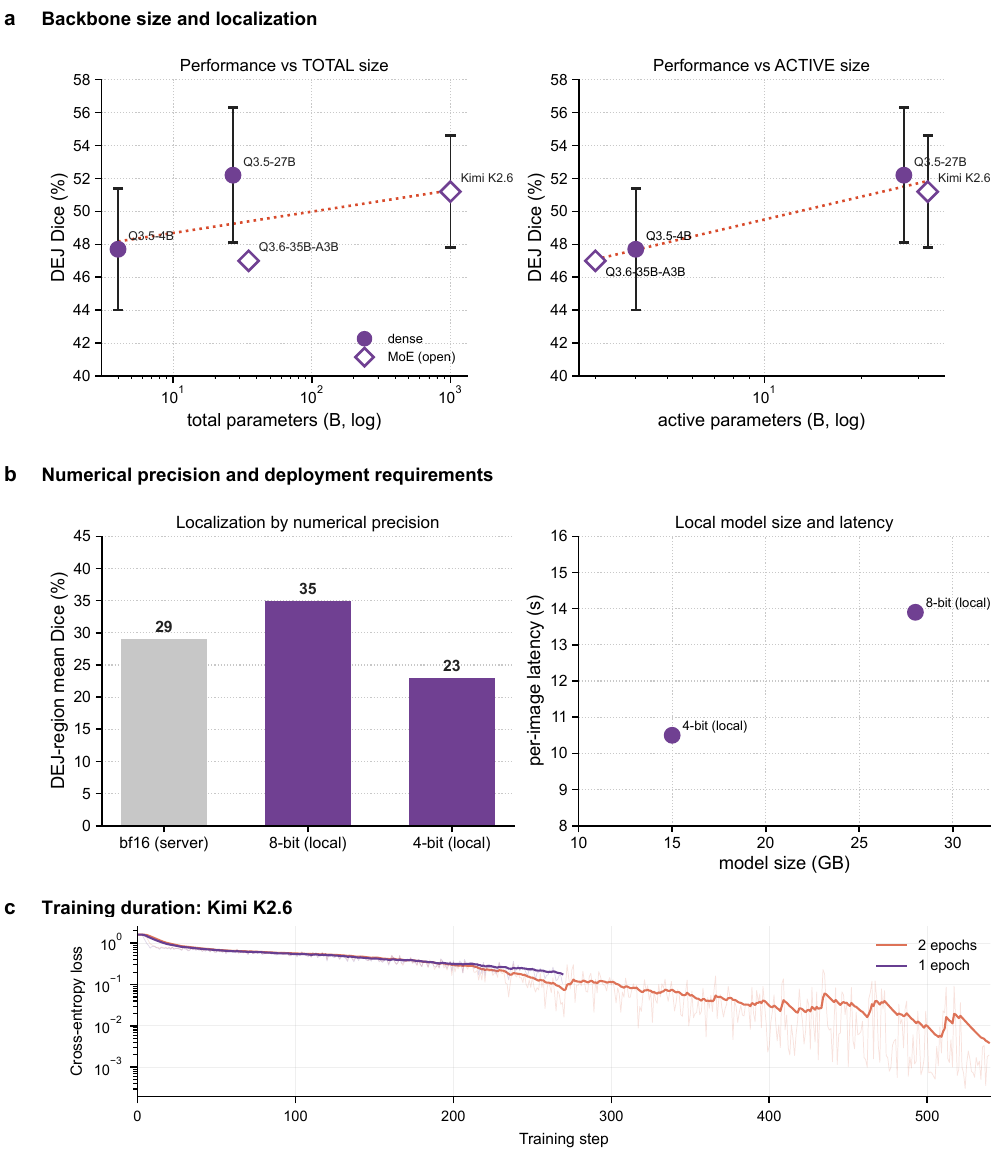}{\textbf{Development studies of model and deployment choices.} \textbf{a,} DEJ localization versus total and active parameter counts, showing the same measurements on two logarithmic axes. Filled markers denote dense models and open markers mixture-of-experts models. \textbf{b,} DEJ-region Dice under bf16 server inference and 8-bit and 4-bit local inference (left); model size and per-image latency for the local configurations (right). \textbf{c,} Kimi K2.6 training cross-entropy over one or two epochs. Thin lines show step-level loss and thick lines its exponential moving average (weight 0.1 on the current step). DEJ, dermoepidermal junction.}{fig:model_choices}}

\expandafter\newcommand\csname suppfigure:fig:organ_supp\endcsname{%
\suppfigure[18cm]{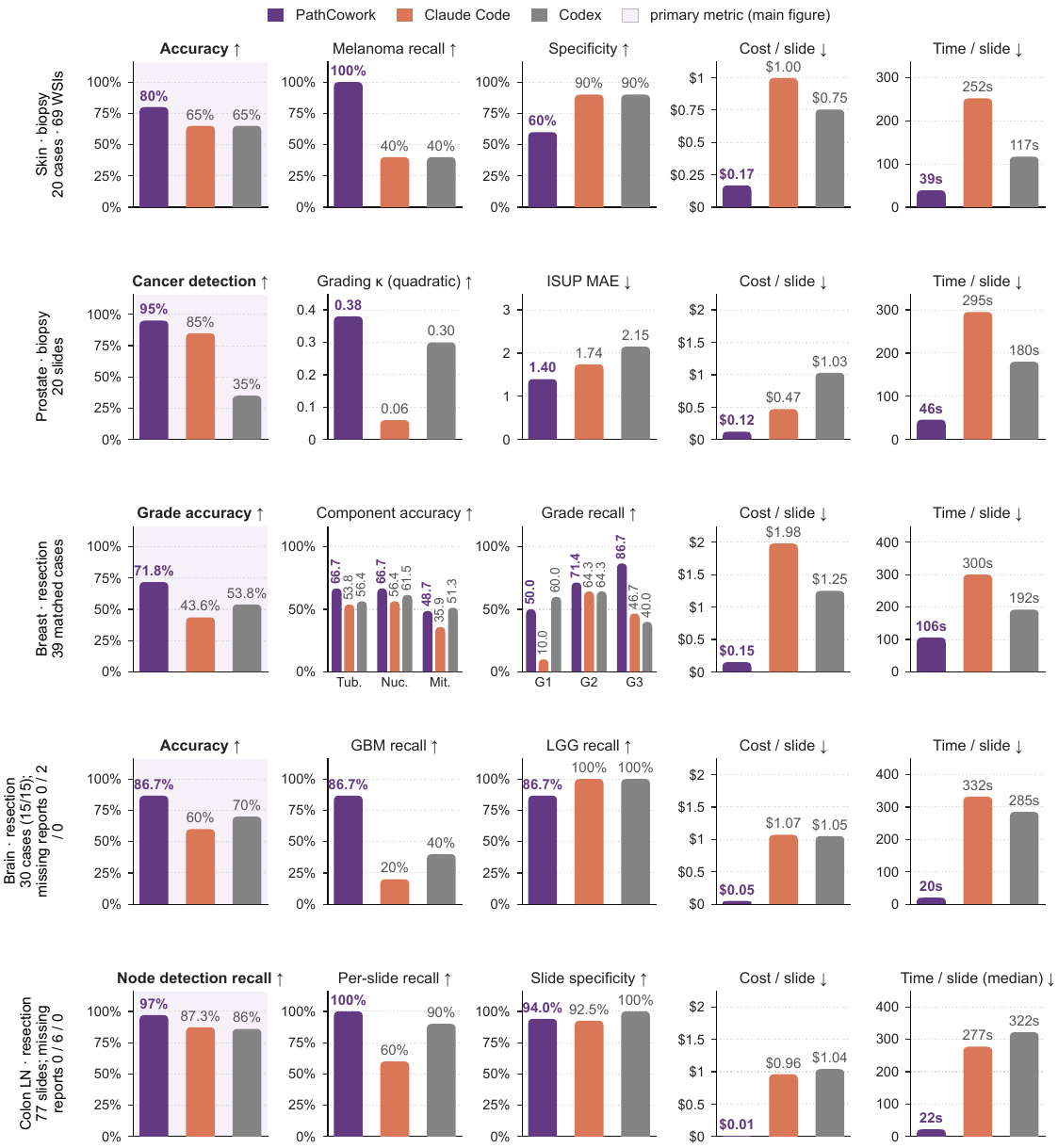}{\textbf{Detailed agent outcomes across five organs.} Rows compare PathCowork (purple), Claude Code (orange) and Codex (grey) on primary and secondary endpoints; shaded panels identify the primary endpoints used in \textbf{Figure~3e}. Colon node detection recall measures the proportion of reference lymph nodes located. In the breast row, Tub., Nuc. and Mit. denote tubule formation, nuclear pleomorphism and mitotic count; G1--G3 denote reference grades. Costs and wall-clock times are per slide; time is the median for colon and the mean for other tasks. Arrows indicate the direction of improvement. Cohorts, model configurations and scoring rules are detailed in Supplementary Note~\ref{supp:agents}.}{fig:organ_supp}}

\expandafter\newcommand\csname suppfigure:fig:skin_grounding_1\endcsname{%
\suppfigure[17cm]{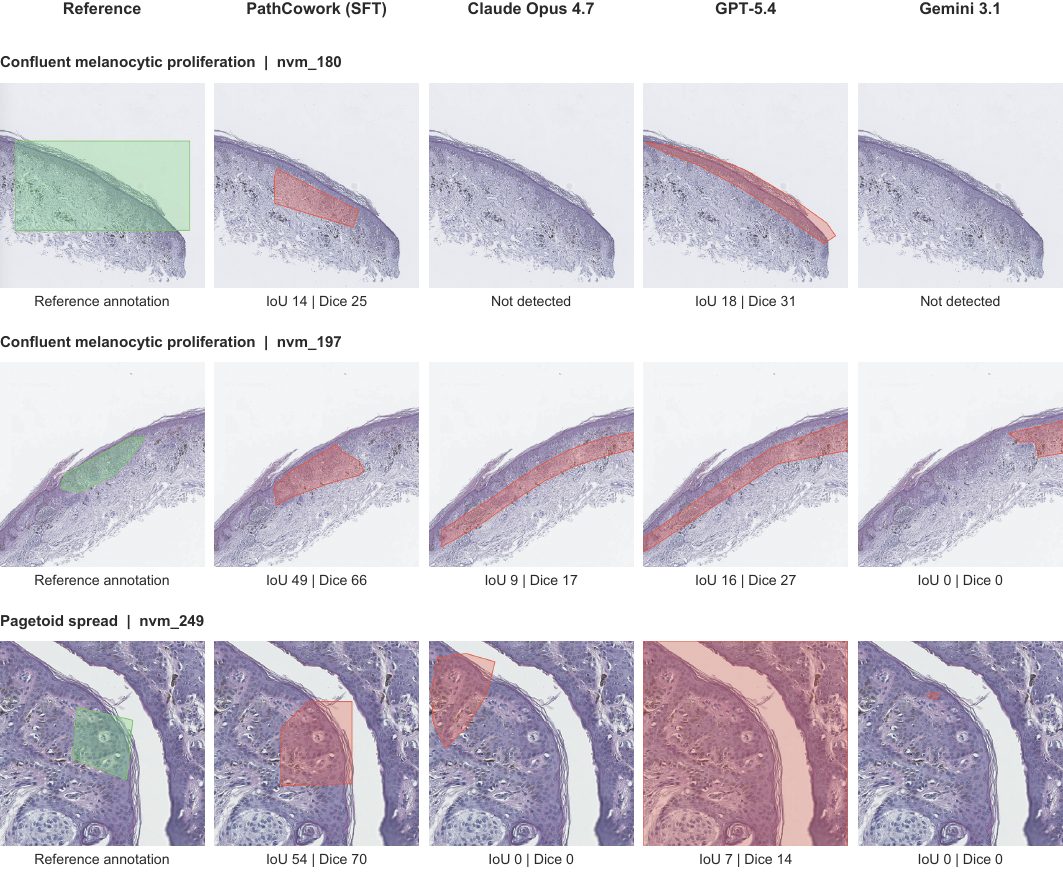}{\textbf{Additional skin grounding examples for confluent growth and pagetoid spread.} Rows show two examples of confluent melanocytic proliferation and one of pagetoid spread. Columns compare the reference annotation with a PathCowork SFT localizer and three frontier models. Reference regions are green and model predictions are red. IoU and Dice are percentages; ``Not detected'' denotes a negative presence prediction.}{fig:skin_grounding_1}}

\expandafter\newcommand\csname suppfigure:fig:skin_grounding_2\endcsname{%
\suppfigure[17cm]{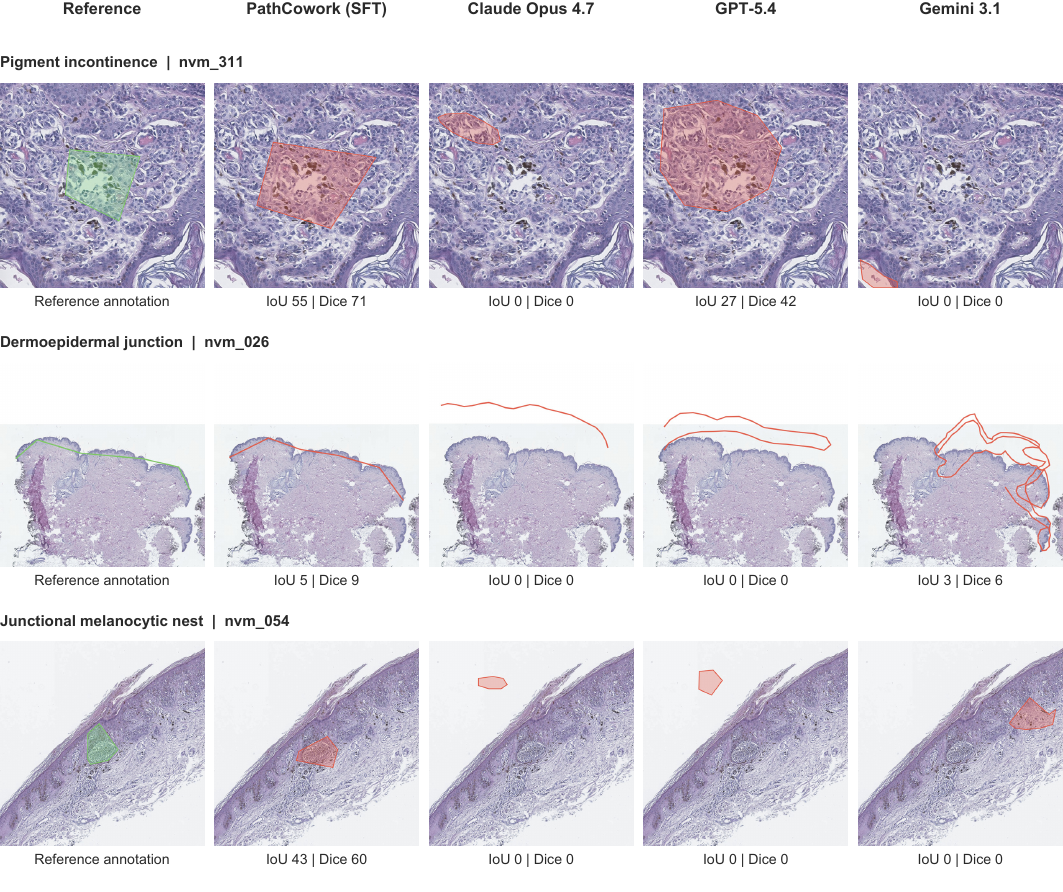}{\textbf{Additional skin grounding examples for pigment, junctional anatomy and nests.} Rows show pigment incontinence, the dermoepidermal junction and a junctional melanocytic nest. Columns, colors and percentage metrics follow \textbf{Supplementary Figure~\ref{fig:skin_grounding_1}}.}{fig:skin_grounding_2}}

\expandafter\newcommand\csname suppfigure:fig:prostate_grounding_1\endcsname{%
\suppfigure[17.6cm]{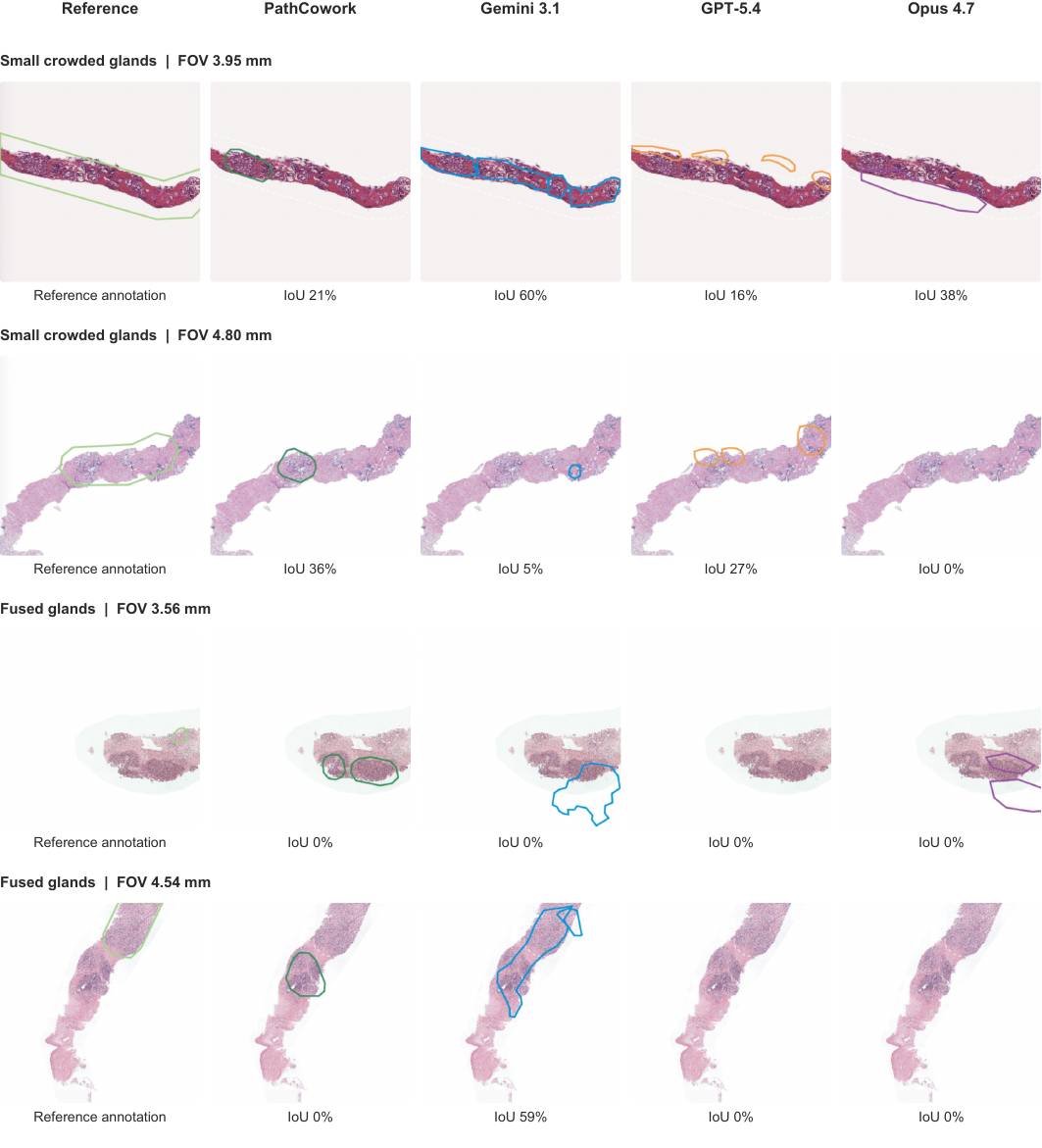}{\textbf{Qualitative grounding comparisons in prostate tissue.} Two examples of small crowded glands and two of fused glands are shown with their reference regions and model outputs. Outlines show reference and predicted regions. Values below each field are example-level IoU percentages. FOV, field of view.}{fig:prostate_grounding_1}}

\expandafter\newcommand\csname suppfigure:fig:prostate_grounding_2\endcsname{%
\suppfigure[17.6cm]{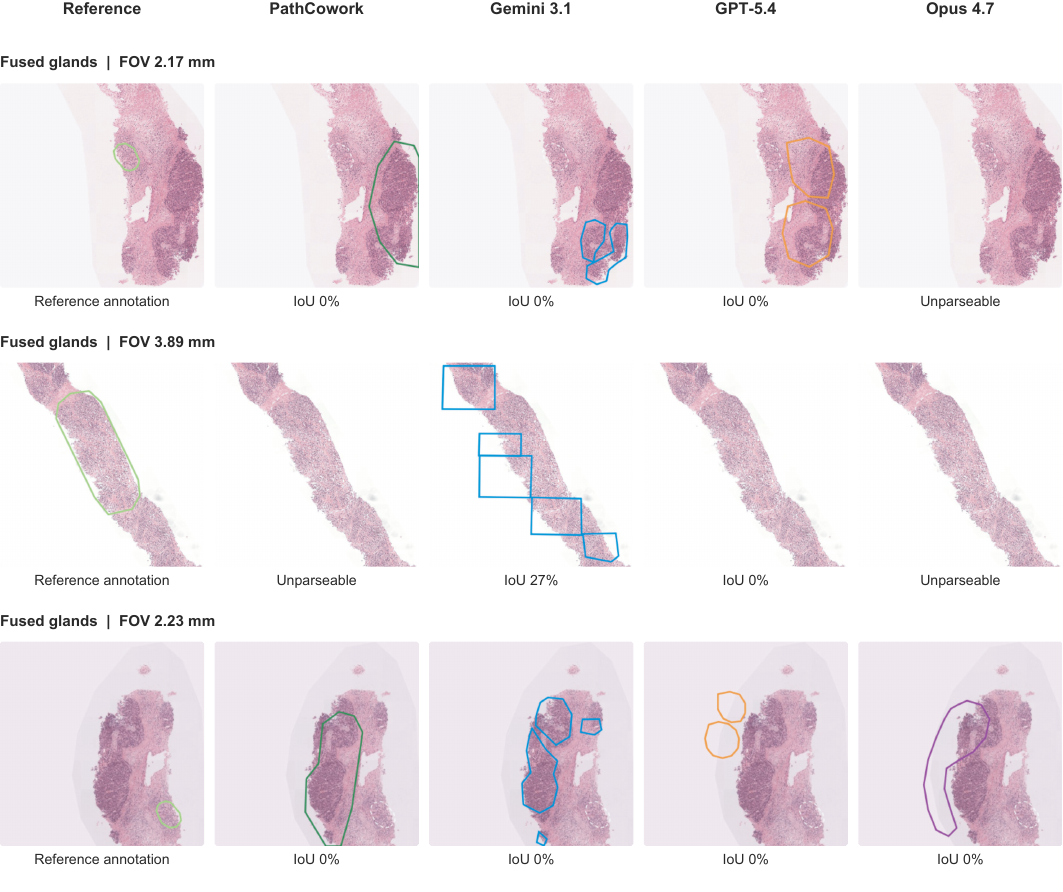}{\textbf{Additional fused-gland grounding examples.} Three further prostate fields are shown with the same column order as \textbf{Supplementary Figure~\ref{fig:prostate_grounding_1}}. Values are example-level IoU percentages. ``Unparseable'' denotes an output without valid spatial geometry.}{fig:prostate_grounding_2}}

\expandafter\newcommand\csname suppfigure:fig:cribriform_examples\endcsname{%
\suppfigure[18cm]{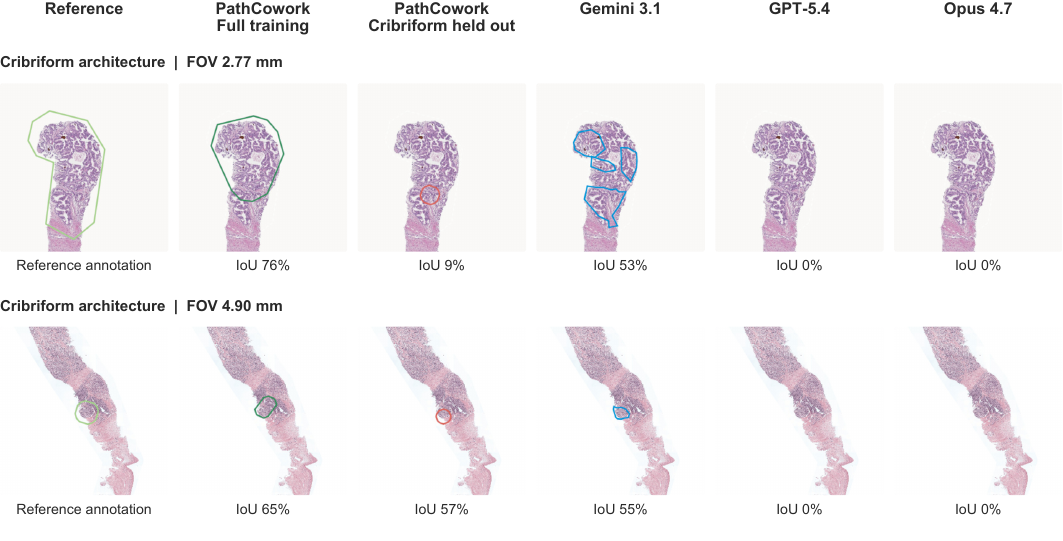}{\textbf{Cribriform localization with and without direct cribriform supervision.} Two prostate fields compare the reference region, PathCowork trained on the full concept set, PathCowork trained with cribriform examples withheld, and three frontier models. Values are example-level IoU percentages. The two fields illustrate variation in transfer to the withheld concept; aggregate results are reported in \textbf{Supplementary Figure~\ref{fig:holdout}}. FOV, field of view.}{fig:cribriform_examples}}

\expandafter\newcommand\csname suppfigure:fig:benchmark_examples\endcsname{%
\suppfigure[14.8cm]{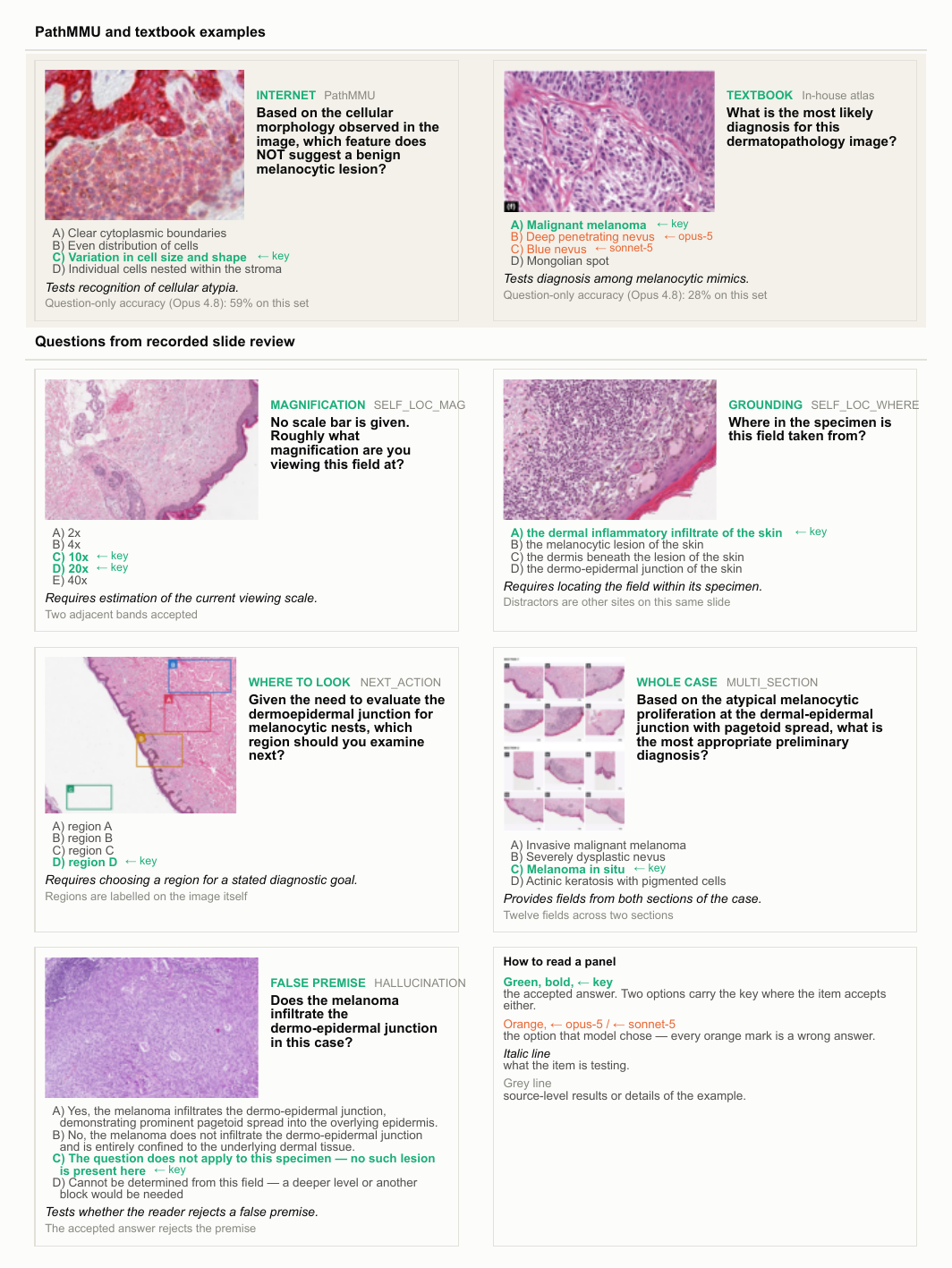}{\textbf{Examples of questions from three benchmark sources.} Top, PathMMU and textbook items. Below, recorded-review items address viewing scale, specimen location, navigation, multiple sections and a false premise; the first two are environmental-sensing subtypes in \textbf{Figure~5}. Green marks accepted answers and orange marks the named models' responses. The magnification example retains the source version's five options and two accepted adjacent bands. Question-only percentages describe source-level Opus~4.8 results.}{fig:benchmark_examples}}

\expandafter\newcommand\csname suppfigure:fig:so1_navigator\endcsname{%
\suppfigure[18cm]{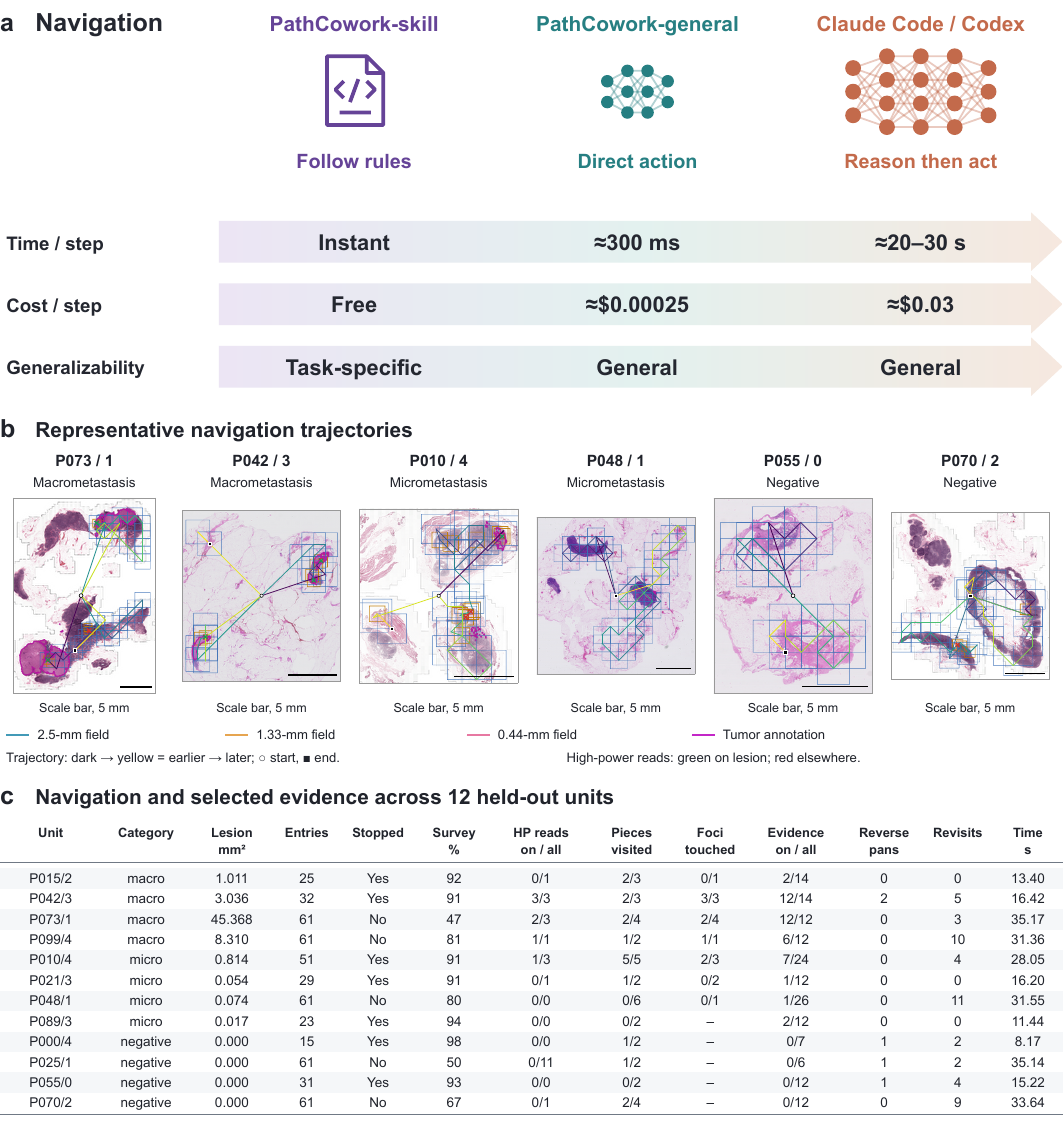}{\textbf{General navigation learned from pathologist behavior.} \textbf{a,} PathCowork-skill follows a compiled examination procedure; PathCowork-general predicts viewing actions from a learned policy; stepwise Claude Code or Codex agents generate an LLM response before acting. Icon sizes are schematic. Time and cost are indicative navigation estimates (Supplementary Note~\ref{supp:so1}). \textbf{b,} Six patient-held-out CAMELYON17 units, showing tissue, annotated tumor, successive fields and high-power reads. Other patients from this dataset contributed training data. Scale bars, 5\,mm. \textbf{c,} All 12 evaluated units, one review each with the supervised navigator, deterministic coverage rules and a nominal 60-step budget. Entries include the terminal observation; budget-limited runs contain 61. Stopped indicates an unforced stop within budget. Survey measures tissue coverage at survey level or higher. High-power (HP) reads count 0.44-mm fields on lesion over all HP fields. Pieces counts fragments of at least 2\,mm$^2$ visited at 1.33\,mm or higher; foci counts annotations of at least 0.02\,mm$^2$ touched at these scales. A dash denotes no eligible focus. Evidence counts retained views overlapping tumor over all retained views. Reverse pans and repeated views are separate counts. Time excludes downstream LLM interpretation.}{fig:so1_navigator}}
 
\expandafter\newcommand\csname suppfigure:fig:recorded_check_coverage\endcsname{%
\suppfigure[16cm]{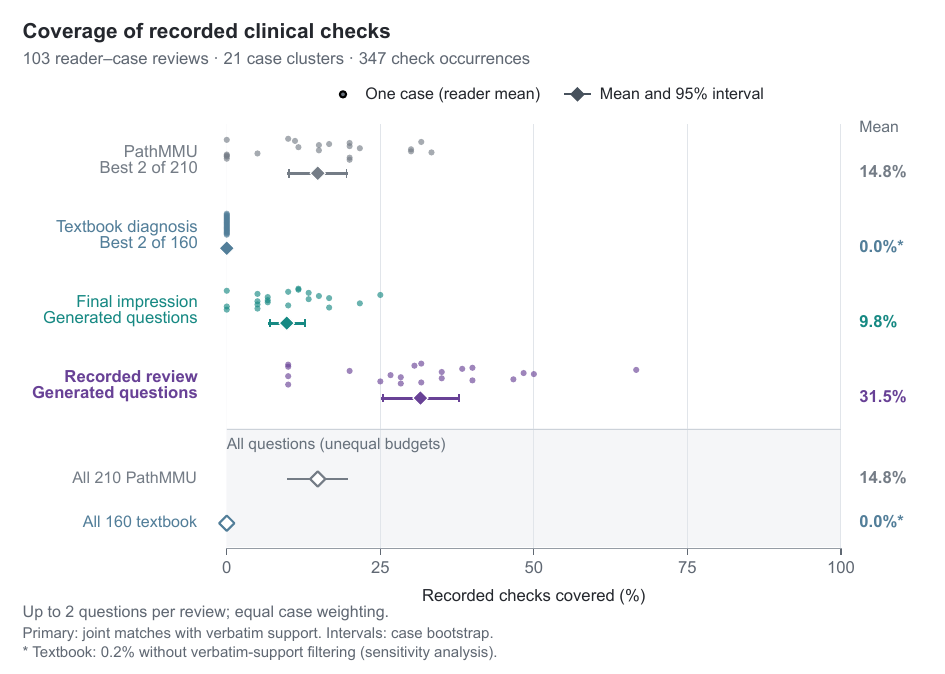}{\textbf{Coverage of recorded clinical checks by four question sources.} For each of 103 reader--case reviews, up to two questions were generated from the recorded narration or a final-impression summary, or retrospectively selected to maximize coverage from the archived PathMMU (210 items) or corrected textbook-derived (160 items) bank. The reference contains 347 selected check occurrences, including similar checks recorded by different readers, across 21 case groups. Coverage requires agreement of two model-based coders on the same check--question match, with verbatim supporting text. Dots show case means after averaging readers; filled diamonds show the equal-case mean and bars the 95\% percentile interval from 10,000 case-bootstrap resamples. Hollow diamonds show full-bank coverage without the two-question limit. The textbook bank consists of atlas-derived diagnosis questions. Asterisk: textbook coverage is 0.2\% in the post-hoc sensitivity analysis without verbatim-support filtering; the same sensitivity rule was applied to all conditions. The analysis measures textual correspondence between recorded checks and question content.}{fig:recorded_check_coverage}}
\suppnote{Additional skin grounding and an ancillary adaptation study}{supp:training}
The main grounding comparisons use supervised fine-tuning of Kimi K2.6 and Qwen3.6-27B on examples derived from the recordings. Each example pairs an image and target concept with a structured spatial response. The dermatopathology grounding evaluation in \textbf{Figure~2e,f} uses the adapted Kimi K2.6 localizer. Additional qualitative skin comparisons from the Qwen SFT development export are shown in \textbf{Supplementary Figures~\ref{fig:skin_grounding_1} and~\ref{fig:skin_grounding_2}}; these examples retain their source model configuration and displayed metrics. Supplementary Table~\ref{tab:model_settings} lists the backbones and configurations used in the additional experiments on data quantity, model size, numerical precision and training duration.

The main test set contains 50 image--concept pairs for each evaluated concept, including 350 pairs across the seven dermatopathology concepts. Training and test patients were separated before generating the examples. Both sets were constructed from behavior-derived candidates with manual spot-checking, augmentation and automatically generated negative examples. Recall uses reference-positive examples as its denominator, precision uses predicted-positive examples, and localization retains missed targets and unparseable outputs with a score of zero.

\suppfig{fig:skin_grounding_1}
\suppfig{fig:skin_grounding_2}

\subsection*{Ancillary development methods}
A separate small-model development study used Qwen3.5-4B~\cite{yang2025qwen3} with rank-8 low-rank adaptation~\cite{hu2022lora} and thinking mode disabled. Its case-level split contained 19 training, three validation and three test cases, corresponding to 936, 148 and 148 message rows. Supervised training used Adam, learning rate $5\times10^{-5}$, batch size 2, maximum sequence length 8,192 tokens and 600 optimization steps. Validation was performed every 50 steps, and the step-450 checkpoint was selected. Selection considered polygon Dice, point-in-target rate, centroid distance, bounding-box intersection over union (IoU) and valid-JSON rate.

This development study also explored an appearance-aware reinforcement-learning objective after supervised training. Recorded attention can identify one instance among several plausible occurrences of a finding in the same field. The reward combined center hit (weight 0.25), normalized centroid proximity (0.20), appearance similarity (0.20), bounding-box IoU (0.15), polygon Dice (0.10) and area agreement (0.10). Appearance similarity was estimated from CIELAB channel statistics and histograms. Training used 25 policy-gradient steps, learning rate $10^{-5}$, four prompts and four sampled responses per prompt, temperature 0.2 and maximum response length 256 tokens. This ancillary SFT-plus-RL experiment is separate from the supervised-only comparisons in the main figures.


\suppnote{Detailed agent outcomes across five organs}{supp:agents}
\textbf{Supplementary Figure~\ref{fig:organ_supp}} reports the organ-specific endpoints underlying the overview in \textbf{Figure~3e}. The evaluations comprise skin biopsies for melanoma versus dysplastic nevus (20 cases, 69 WSIs), prostate needle biopsies for cancer detection and ISUP grading (20 slides), breast resections for Nottingham grading (39 matched cases), brain resections for glioblastoma versus lower-grade glioma (30 cases, 15 per class) and colon-resection lymph nodes for metastasis detection (77 slides, 10 positive and 67 negative).

Brain cases were drawn from TCGA-GBM and TCGA-LGG~\cite{tcga2013pancancer}, breast and melanocytic cases from HISTAI~\cite{nechaev2025histai}, prostate slides from PANDA~\cite{bulten2022panda}, and colorectal lymph-node slides from LNCO2~\cite{maras2020lnco2} (Supplementary Table~\ref{tab:agent_cohorts}). Evaluation cases were unseen during grounding training and skill construction. PANDA ISUP labels supplied the prostate grading reference, and LNCO2 annotations supplied the metastatic-node references.

The organ-specific experiments were conducted at different times using the strongest models then available in each agent framework. The prostate comparators used Claude Code with Sonnet and Codex with GPT-5.5; their evaluation runner allowed up to ten image crops and a 480-s wall-clock limit per slide. Codex used GPT-5.6-sol for brain and colon, with high reasoning effort for colon. The breast Claude Code comparator used Sonnet 5, and the brain PathCowork run used v4.4 High. PathCowork completed the prescribed skill steps without a task-level timeout, as described in Methods.

In skin, PathCowork achieved melanoma recall of 100\% and specificity of 60\%, compared with 40\% recall and 90\% specificity for both general-purpose agents. Its higher overall accuracy therefore accompanied lower specificity. In prostate, quadratic-weighted ISUP $\kappa$ was 0.38, 0.06 and 0.30 for PathCowork, Claude Code and Codex, respectively; the corresponding grade mean absolute errors were 1.40, 1.74 and 2.15. Prostate detection uses all 20 slides, with the Claude Code timeout counted as a failure (17/20, 85\%); its grading $\kappa$ and MAE retain the 19 valid predictions.

The brain comparison uses 30 cases (15 GBM and 15 LGG) for every agent. PathCowork correctly classified 13/15 GBM and 13/15 LGG cases (26/30 overall); Claude Code classified 3/15 and 15/15 correctly (18/30); Codex classified 6/15 and 15/15 correctly (21/30). Missing reports numbered zero, two and zero, respectively, and remain in the denominator as incorrect outcomes. Mean times were 20, 332 and 285 seconds per slide, respectively.

The breast comparison uses one matched set of 39 cases for all agents. In the order PathCowork, Claude Code and Codex, overall grade accuracy was 28/39 (71.8\%), 17/39 (43.6\%) and 21/39 (53.8\%). Correct component scores were 26, 21 and 22 for tubule formation; 26, 22 and 24 for nuclear pleomorphism; and 19, 14 and 20 for mitotic count, all out of 39. Grade-specific recall was 5/10, 1/10 and 6/10 for grade 1; 10/14, 9/14 and 9/14 for grade 2; and 13/15, 7/15 and 6/15 for grade 3. PathCowork's largest grade-specific advantage was therefore in grade 3, while Codex had higher grade 1 recall and correctly scored one more case for mitotic count. Per-slide costs were \$0.15, \$1.98 and \$1.25, and times were 106, 300 and 192\,s, respectively (\textbf{Supplementary Figure~\ref{fig:organ_supp}}).

For LNCO2, Claude Code achieved 87.3\% node detection recall, called 12/15 ground-truth tumour nodes positive (80\% end-to-end per-node metastasis recall), and called 101/101 non-tumour nodes negative (100\% per-node specificity). Missed tumour nodes remain in the 15-node denominator as false negatives. Missing reports numbered zero, six and zero for PathCowork, Claude Code and Codex, respectively. Claude Code's slide-level recall was 6/10 (60\%) and specificity was 62/67 (92.5\%); its two missing positive-slide reports and four missing negative-slide reports counted as failures. Codex achieved slide-level recall of 9/10 (90\%) and specificity of 67/67 (100\%).

Costs are reported per slide at the recorded API prices; the Codex colon cost was \$1.04 per slide. Times are wall-clock values per slide, summarized by the median for colon and the mean for other tasks.

\suppfig{fig:organ_supp}

\clearpage
\suppnote{General navigation learned from pathologist behavior}{supp:so1}

The five agents in \textbf{Figure~3} turn recorded case experience into procedures for recurring tasks. Open-ended questions can require findings or relationships beyond this repertoire, including examinations only partly represented in the available recordings.

PathCowork-general learns reusable viewing actions---surveying tissue, moving between fields, changing magnification and returning to a wider context---to gather evidence beyond a predefined skill. The proposed workflow assigns navigation to this learned policy and interpretation to an LLM, extending the use of recorded behavior to questions for which a complete procedure has yet to be compiled. The experiments below evaluate navigation and evidence collection; combined question answering is a future application.

\paragraph{Two complementary uses of recorded behavior.}
\href{https://github.com/zhihuanglab/PathCowork/blob/main/videos/03_Supplementary_Video_3_General_Navigation.mp4}{\textbf{Supplementary Video 3}} shows a recorded example of the learned navigator examining a lymph-node slide.
We refer to the task-specific approach in \textbf{Figure~3} as PathCowork-skill and the learned navigation approach as PathCowork-general (\textbf{Supplementary Figure~\ref{fig:so1_navigator}a}). The former encodes examination steps explicitly; the latter learns successive viewing decisions from behavioral records and reference trajectories. Both reuse pathologist experience. A stepwise agent such as Claude Code or Codex instead invokes the LLM to choose successive viewing actions. Here, ``general'' denotes a reusable navigation policy shared across tasks.

\paragraph{Indicative navigation time and cost.}
The indicative values in \textbf{Supplementary Figure~\ref{fig:so1_navigator}a} cover navigation decisions. ``Instant'' and ``Free'' denote script overhead. The learned-policy estimate of 300\,ms is based on a 318-ms forward pass of an earlier 384-pixel, 20-slot prototype on an M5 Pro with MLX bf16. Its hosted-inference equivalent is approximately US\$0.00025 per decision: 753 image-plus-prompt tokens at the \href{https://tinker-docs.thinkingmachines.ai/tinker/models/}{Tinker 4B prefill rate} of US\$0.33 per million tokens, accessed 21 September 2026. LLM estimates are 20--30\,s and approximately US\$0.03 per step, based on US\$1 per 30 steps. These estimates cover decision computation, excluding image loading, downstream interpretation, training and hardware costs.

\paragraph{From recorded examination to navigation targets.}
The general navigator combines processed pathologist gaze and viewport records with trajectories generated by a reference policy on annotated slides. Both sources are mapped to the same viewing scales and action vocabulary. Human records preserve the operations taken during examination; reference trajectories supply broader image coverage and examples of surveying tissue, approaching an annotated focus and returning to continue the survey. Perturbed reference trajectories provide corrective targets in states outside the usual path, motivated by the distribution-shift problem addressed by dataset aggregation for imitation learning~\cite{ross2011dagger}. Images and spatial targets are transformed together under rotations and reflections. Before subsampling for training, the mixed-supervision pool contained 40,000 records, including 2,021 from human recordings.

\paragraph{An exploratory comparison of behavioral supervision.}
An earlier model comparison added 1,916 human navigation records to the same 5,000 reference-navigation and 6,000 perception records used by a control. Both models used one training epoch and the same training recipe, although adding records also increased the number of optimization steps. On 286 held-out views with a defined human action, agreement with that action increased from 24.8\% to 60.1\%. Among the 181 views followed by a human pan, the control predicted pan zero times and the model with human records predicted it 105 times. Human records increased use of the pan operation, although agreement remained below the 63.3\% majority-action baseline and repetitive panning persisted. Each condition used one run of an earlier model and state representation.

\paragraph{A local model for successive viewing decisions.}
The general navigator uses a four-level ladder: the tissue overview followed by square fields of 4.0, 1.33 and 0.44\,mm; the survey field is 2.5\,mm for lymph-node slides. Each field is rendered at $384\times384$ pixels and divided into a $4\times4$ grid. The state records viewing scale, remaining steps, tissue and coverage fractions, areas already examined at a deeper level, directions containing unsurveyed tissue and the last three actions. Per-cell survey coverage is computed before marking the current arrival, allowing a new field to be distinguished from a revisit.

The navigator selects among zooming into a grid cell, zooming out, panning in one of eight directions and stopping. Its general-model template contains 29 answer slots: 16 cell-level attention outputs, three action outputs, nine field-description outputs and one unused deferral output. A single forward pass supplies logits at fixed answer positions; softmax over the allowed answer letters gives each slot's distribution, without autoregressive text generation. This use of predefined decision outputs follows the System One design principle described for Jev~\cite{typesafe2026jev}; our implementation adapts a vision--language model to pathology images and viewing behavior. It uses Qwen3.5-4B~\cite{yang2025qwen3} with rank-32 LoRA~\cite{hu2022lora}. The evaluated general model was trained through Tinker~\cite{thinkingmachines2025tinker} for one epoch on 16,000 navigation and 12,000 perception records, with batch size 32 and peak learning rate $2\times10^{-4}$. Cross-entropy is restricted to defined answer positions; sampling an answer from a soft target distribution matches that target in expectation. The best validation checkpoint was the end-of-epoch checkpoint. Local inference used merged bf16 weights with MLX.

\paragraph{Coverage bookkeeping and evidence selection.}
Deterministic rules constrain the model's choices: they restrict repeated entry into already examined cells, veto reverse pans when another direction contains unsurveyed tissue, and return to a wider view when continued panning adds no coverage. At the end of a review, distinct non-overview views are grouped by tissue fragment, and up to six per fragment are retained according to the maximum of their predicted tumor, interface and necrosis scores. Survey views remain eligible to preserve lower-power architecture. The retained images form the evidence set for subsequent interpretation.

\paragraph{Evaluation on held-out lymph-node units.}
The general-model evaluation used 12 CAMELYON17 lymph-node units~\cite{bandi2019camelyon17}: four labelled macrometastasis, four micrometastasis and four negative. Patients were held out from the CAMELYON17 training sources; training images included other patients from the same dataset. On 400 navigation records, the 6,400 cell outputs achieved AUROC 0.965 and average precision 0.455, compared with 0.886 and 0.101 when the image was replaced by a white field while retaining the state. Operation agreement was 88.2\% versus 86.8\% in that control; pan-direction agreement was 41.7\% versus 35.4\% over 223 pan-labelled records. The image contribution varied by output and scale: for overview-level cell outputs, average precision was 0.108 with the image and 0.186 without it.

\paragraph{What the resulting reviews captured and missed.}
Across one review per unit, median tissue survey coverage was 91\%, and seven of 12 reviews stopped within the nominal 60-step budget (\textbf{Supplementary Figure~\ref{fig:so1_navigator}c}). Four of eight metastatic units received both a mid-power visit overlapping an annotated focus and a high-power read on the lesion. The selected evidence set contained a view overlapping the annotation in all eight, including survey views from the four units without a high-power lesion read. Representative successes and failures are shown in \textbf{Supplementary Figure~\ref{fig:so1_navigator}b}. Eight of 12 reviews had no immediate reverse pan; five reached the budget, and two had no mid-power visit to any qualifying tissue fragment. Recorded local review times ranged from 8.17 to 35.17\,s, excluding downstream LLM interpretation. These exploratory results identify coverage, revisiting and stopping as unresolved behavioral problems. The navigator weights are released at \url{https://github.com/zhihuanglab/PathCowork}; its training corpus is not included in the release.

\suppfig{fig:so1_navigator}
 
\suppnote{Frozen tile selection on the matched prostate cohort}{supp:eagle}
We compared behavior-derived spatial supervision with EAGLE, an alternative that selects tissue using attention from a pretrained model~\cite{neidlinger2026eagle}. The comparison used the fixed 30-slide PANDA test cohort from \textbf{Figure~4}, containing five slides in each ISUP grade group from 0 to 5. Both EAGLE resolutions used exactly the same 20 training subsets as the $N=20$ CoworkMIL evaluation, sampled from the 65-slide training pool.

For each resolution, frozen CTransPath features~\cite{wang2022ctranspath} were ranked by frozen CHIEF attention~\cite{wang2024chief}, and up to 25 selected tiles were re-encoded with Virchow2~\cite{zimmermann2024virchow2}. The mean of their class-token embeddings formed the slide representation. A logistic-regression classifier was fitted to the 20 training slides, with the encoders and tile selector frozen. EAGLE achieved mean quadratic-weighted $\kappa$ of 0.430 at 0.5\,$\mu$m per pixel and 0.435 at 2.0\,$\mu$m per pixel, with mean absolute grade errors of 1.225 and 1.298, respectively. On these same training subsets and test slides, Virchow2-CoworkMIL achieved $\kappa=0.486$ and mean absolute error 1.178, compared with 0.385 and 1.347 for Virchow2-ABMIL (Supplementary Table~\ref{tab:eagle_matched}).

EAGLE uses 1,280-dimensional class-token features and mean pooling, whereas the Virchow2 MIL models use 2,560-dimensional concatenated class-token and mean patch-token features with a trained aggregator. Spatial sampling also differs, and CoworkMIL additionally receives grounding-derived supervision during training. Statistical summaries follow the fixed-cohort analysis described in Methods.

\suppnote{Benchmark sources and coverage of recorded clinical checks}{supp:outputs}
\textbf{Supplementary Figure~\ref{fig:benchmark_examples}} compares illustrative questions from PathMMU~\cite{sun2024pathmmu}, a textbook atlas and recorded review.

Generated questions and accepted-answer keys underwent manual review using the source images and recorded context. The retained question-bank composition is given in Supplementary Table~\ref{tab:benchmark_composition}; environmental sensing combines viewing-scale and tissue-location questions. The separate source comparison uses 100 questions from each source, and the paired Opus image-input comparison uses 148 recorded-review questions, as described in Methods.

\suppfig{fig:benchmark_examples}

\subsection*{Coverage of recorded clinical checks}

We asked whether questions constructed from recorded review retained checking content that was less readily represented by a final impression or existing question banks. The reference contained 347 selected check occurrences from 103 reader--case reviews across 21 case groups; checks expressed by different readers were retained separately. With up to two questions per review, narration-derived questions covered 31.5\% of the reference, compared with 9.8\% for questions generated by the same model from the reader's final impression. Retrospective selection of the best-matching pair yielded 14.8\% coverage from the 210-item PathMMU dermatopathology bank and 0.0\% from the corrected 160-item textbook-derived bank (\textbf{Supplementary Figure~\ref{fig:recorded_check_coverage}}). Narration-derived coverage was 16.7 percentage points higher than PathMMU (95\% case-bootstrap interval, 8.5--24.8) and 31.5 points higher than the textbook bank (25.4--37.8).

Using all questions in each bank gave 14.8\% coverage for PathMMU and 0.0\% for the textbook bank. Restricting the latter to the corrected 100-item subset used in the main source comparison gave 0.0\%. No textbook match passed both joint coding and verbatim-support validation. The evaluated bank contained diagnosis-only questions; identifying a diagnosis was not credited as testing every observation or examination action that could inform it. Separately, the two coders assigned textbook best-pair coverage of 1.3\% and 14.4\%, respectively; the primary analysis required their agreement on the same requirement--question match.

Recorded content included examination plans as well as morphological observations. A search for lymphovascular invasion and a proposed stain to exclude occult invasion were retained in narration-derived questions. Generation remained incomplete: a recorded pigment check was omitted by both generated question sets. The comparison measures retention of recorded checking content under a limited question budget.

The zero textbook estimate was sensitive to the quotation check: both coders matched a scar question, but one supplied a composite description of its options rather than a verbatim supporting phrase. In a post-hoc analysis retaining all in-scope model-proposed matches without requiring verbatim supporting phrases, coverage was 34.4\% for recorded review, 11.3\% for final impressions, 15.3\% for PathMMU and 0.2\% for the textbook bank. This sensitivity rule was applied to all conditions without new inference; the ordering was unchanged.

\suppfig{fig:recorded_check_coverage}

\subsection*{Content-coverage methods}

This content-coverage experiment was separate from the image-supported benchmark construction described in the main Methods. Five readers examined the same 30 whole-slide images, grouped under 21 case prefixes. Companion sections and repeat visits were pooled within each reader--case combination. Before question generation, a language-model assistant extracted explicit observations, checking actions and requests for additional information from the transcripts, retaining source quotations and timestamps. Selection across recognition, characterization, examination completion and additional-information categories retained at most four checks per combination. Repeated checks within a combination were merged, while similar content expressed by different readers remained separate. Two combinations without an explicit check were excluded, leaving 103 evaluable reviews and 347 check occurrences. The resulting reference was a capped set of assistant-extracted checks.

Claude Sonnet 5 generated up to two four-option questions from either the cleaned narration or a source-supported summary of the same reader's final impression, using identical instructions and allowing abstention. The impression summaries were constructed from the transcripts for this experiment and retained uncertainty and associated findings. Generation used text-only inputs, and the resulting candidate questions were retained as generated.

The comparison banks contained 210 archived PathMMU dermatopathology questions~\cite{sun2024pathmmu} and 160 questions constructed from \textit{Dermatopathology: Diagnosis by First Impression}~\cite{ko2022dermatopathology}. The latter were locally assembled diagnosis questions. We used corrected intended answers linked to the printed atlas panel letters and preserved the original stems and choices. All 160 stems asked for the most likely diagnosis. The corrected 100-item subset used in the main source comparison was analyzed separately. Bank sizes and question formats therefore differed.

GPT-5.6-Sol and Claude Haiku 4.5 independently matched checks to candidate questions. Source labels and reader identities were withheld; items had opaque identifiers. Intended answers distinguished the central tested subject from distractor mentions. The rubric required correspondence to both the clinical target and the checking operation. For example, identifying invasion did not automatically cover deciding to obtain an additional stain. Each proposed match included a literal phrase checked against the question text. Primary coverage required both coders to accept the same check--question match and provide valid verbatim support; individual-coder results were retained as sensitivity analyses. Generation and matching used subscription CLIs, without tools. The textbook extension reused the frozen rubric and settings in separate textbook-only packets. Original-condition matches were retained, so packet context and evaluation occasion differed between the original and added conditions.

For each review, exhaustive retrospective selection identified up to two bank questions maximizing the union of covered checks. Selection used the recorded checks as the retrieval reference. Full-bank unions summarized coverage using every question in each bank. Coverage proportions were averaged equally across readers within case, then equally across the 21 cases. Paired differences and mean intervals used 10,000 case-bootstrap resamples (seed 20260920). Narration supplied both the reference and the richer generation input, making within-record content retention the evaluation target.
 
\suppnote{Data scaling and concept transfer}{supp:scaling}
We examined training-data dependence in a prostate low-power triage task using a Qwen3.5-27B localizer. The target vocabulary contained eight architectural patterns: small crowded glands, infiltrative growth, glands out of place, fused or poorly formed glands, cribriform architecture, glomeruloid architecture, sheets or single cells, and comedonecrosis. Training sets contained 10, 20, 30 or 51 annotated WSIs, with two epochs at each size.

Detection recall increased from approximately 61\% at 10 WSIs to 90\% at 51 WSIs, while localization IoU remained between approximately 41\% and 46\% (\textbf{Supplementary Figure~\ref{fig:scaling_mech}}). All eight concepts were represented in the smallest set, and the number of training polygons increased from 752 to 4,024. Additional examples primarily improved detection coverage within this concept vocabulary.

\suppfig{fig:scaling_mech}

The per-concept plots show this distinction for six architectural patterns (\textbf{Supplementary Figure~\ref{fig:perconcept}}). For example, recall for infiltrative growth increased from approximately 36\% to 100\%, and recall for fused or poorly formed glands from approximately 38\% to 88\%, across the plotted training sets. Changes in IoU were smaller and varied in direction across concepts.

\suppfig{fig:perconcept}

Training budget also affected the comparison. Under a fixed budget of 100 optimization steps, parse rate declined from approximately 97\% to 69\% as the dataset increased; the largest fixed-step run received about 0.28 epochs. Parse rate remained near 100\% in the two-epoch series (\textbf{Supplementary Figure~\ref{fig:scaling_compute}}). The fixed-step runs hold the optimization budget constant; the equal-epoch runs expose each training example a comparable number of times.

\suppfig{fig:scaling_compute}

The adapted model achieved higher IoU than the displayed frontier baselines on each of the six plotted concepts (\textbf{Supplementary Figure~\ref{fig:baseline}}). Individual prostate fields show substantial variation, including non-overlapping predictions and unparseable responses (\textbf{Supplementary Figures~\ref{fig:prostate_grounding_1} and~\ref{fig:prostate_grounding_2}}).

\suppfig{fig:baseline}
\suppfig{fig:prostate_grounding_1}
\suppfig{fig:prostate_grounding_2}
 In a separate experiment withholding cribriform supervision, cribriform IoU was 29\%, compared with 54\% after full concept training, 42\% for Gemini 3.1 and 0\% for GPT-5.4 and Opus 4.7 (\textbf{Supplementary Figure~\ref{fig:holdout}}). The retained performance indicates some transfer to this withheld concept, while the difference from full training shows the additional value of concept-specific examples. Two qualitative comparisons illustrate this variation at the level of individual fields (\textbf{Supplementary Figure~\ref{fig:cribriform_examples}}).

\suppfig{fig:holdout}
\suppfig{fig:cribriform_examples}

\suppnote{Model and deployment choices}{supp:backbone}
Complementary development experiments examined the trade-offs underlying model and deployment choices: localization across model sizes, resource use at reduced precision and training behavior over one or two epochs. The settings for each experiment are given below.

Backbone comparisons assessed dermoepidermal-junction localization using Dice overlap. The four-model comparison includes Qwen3.5-4B, Qwen3.6-35B-A3B, Qwen3.5-27B and Kimi K2.6 (\textbf{Supplementary Figure~\ref{fig:model_choices}a}). Mean Dice was approximately 47--48\% for the smaller dense and mixture-of-experts models and 51--52\% for Qwen3.5-27B and Kimi K2.6. The same measurements are plotted against total and active parameter counts to compare these dense and mixture-of-experts backbones; architecture and parameter count vary together.

A separate 27B dermatopathology experiment compared bf16 server inference with 8-bit and 4-bit local inference (\textbf{Supplementary Figure~\ref{fig:model_choices}b}). Mean dermoepidermal-junction-region Dice was 29\%, 35\% and 23\%, respectively. The local 4-bit model occupied approximately 15\,GB, compared with 28\,GB for 8-bit, and per-image latency decreased from approximately 14 to 10.5\,s. Both numerical precision and the execution environment differed across these configurations.

Training loss for Kimi K2.6 continued to decrease during a second epoch, reaching the order of $10^{-3}$ near the end of the run (\textbf{Supplementary Figure~\ref{fig:model_choices}c}). The one-epoch trace ended at a higher training loss.

\suppfig{fig:model_choices}

\suppnote{Spatial formats and reinforcement learning}{supp:recipe}

\subsection*{Choosing an output format before constructing training targets}

The six-format development comparison tested whether Qwen3.5-4B could express a usable spatial response before supervised training. Prompts requested box tokens, a plain polygon, JSON polygons with one hypothetical example, the same JSON format without the example, a clockwise boundary trace, or a segmentation schema containing a box, mask and label. The common 60-tile set was drawn from Ivy GAP~\cite{allen2015ivygap,puchalski2018ivygap}; selection criteria are described in Methods. Output geometry, region purity and coverage were assessed separately.

The box-token and plain-polygon prompts produced 50 and 53 regions, respectively, all classified as rectangles. Adding one hypothetical JSON example elicited regions on 37 of 60 tiles, with 28 of 37 regions classified as non-rectangular and presence accuracy of 0.75. Removing the example reduced region-producing responses to 5 of 60 tiles and presence accuracy to 0.38 in the archival rerun. Without the example, purity increased from 0.17 to 0.41 and coverage was 0.01. The boundary-trace and segmentation-schema prompts yielded non-rectangular regions in 8 of 26 and 7 of 69 predictions, respectively. We selected JSON with one example because it combined frequent region output, preserved presence discrimination and predominantly non-rectangular geometry.

Conditions 1--3 use preserved original predictions; conditions 4--6 use reruns from 6 September 2026 with reconstructed prompts archived alongside the inputs. The selected JSON example illustrated the response structure for a hypothetical image.

\subsection*{Effect of supervised initialization}
Three atlas-derived glioma experiments compared reinforcement learning from the base model, from a supervised checkpoint, and from a supervised checkpoint trained with additional lower-grade glioma (LGG) negatives. Each condition used one training run; the protocol and reward are specified in Methods.

The base and supervised-initialized runs used the same 2,553-record pool. LGG augmentation affected both supervised initialization and the reinforcement-learning pool; the corresponding supervised validation sets contained 77 records before augmentation and 117 afterward.

Between the first and final validation checks, reward decreased from 0.478 to 0.306 and presence accuracy from 0.75 to 0.58 when reinforcement learning began from the base model. With supervised initialization, validation reward increased from 0.512 to 0.573 and presence accuracy remained near 0.83--0.84. The run with additional LGG negatives reached a validation reward of 0.665 and presence accuracy of approximately 0.89. These results motivated supervised initialization and joint monitoring of localization and presence discrimination.

Supervised logs track response-token negative log-likelihood on training and validation sets. Reinforcement-learning logs track mean reward from 12 prompts per step, alongside greedy-validation reward and presence accuracy at four evaluation events.
 \clearpage
\section*{Supplementary Tables}
\pdfbookmark[\suppchildbookmarklevel]{Supplementary Tables}{supp.tables}
\begin{table}[H]
\centering
\caption{\textbf{Datasets represented in the recording collection.} Image sources include TCGA~\cite{tcga2013pancancer}, DROID-skin~\cite{lindman2019droidskin,stadler2021database} and LNCO2~\cite{maras2020lnco2}; behavioral recordings were acquired through the study platform.}
\label{tab:datasets}
\small
\renewcommand{\arraystretch}{1.18}
\begin{tabular}{L{0.23\linewidth}L{0.48\linewidth}L{0.18\linewidth}}
\toprule
Dataset & Tissue or subspecialty & Source type \\
\midrule
TCGA-COAD & Colorectal pathology & Public \\
TCGA-STAD & Gastric pathology & Public \\
TCGA-BLCA & Bladder pathology & Public \\
TCGA-PRAD & Prostate pathology & Public \\
TCGA-DLBC & Hematopathology & Public \\
TCGA-GBM & Neuropathology & Public \\
DROID-skin & Dermatopathology & Public \\
LNCO2 & Colorectal lymph-node pathology & Public \\
Cytology teaching set & Cytopathology & Institutional \\
\bottomrule
\end{tabular}
\end{table}

\begin{table}[H]
\centering
\caption{\textbf{Intermediate record fields.} Coordinates and timestamps link derived examples to the original slide and recording.}
\label{tab:data_schema}
\small
\renewcommand{\arraystretch}{1.18}
\begin{tabular}{L{0.17\linewidth}L{0.24\linewidth}L{0.48\linewidth}}
\toprule
Field & Representation & Content \\
\midrule
Image & Crop and WSI metadata & Viewed tissue region, slide identifier, coordinates and magnification \\
Narration & Text and timestamps & Transcribed utterance, neighboring speech and its time interval \\
Grounding & Spatial coordinates & Polygon, bounding box or polyline associated with a narrated concept \\
Trajectory & Ordered event records & Sequence of views, positions, magnifications and transitions \\
Revisit & Linked event records & Previous observations of the same WSI region \\
Derived context & Structured text & Normalized description, case summary and interpretation of the narration \\
Video & Session recording & Visual reference for quality review and event alignment \\
\bottomrule
\end{tabular}
\end{table}
\clearpage

\begin{table}[H]
\centering
\caption{\textbf{Model configurations by analysis.} Backbone identifiers and training conditions are listed for the individual experiments.}
\label{tab:model_settings}
\small
\renewcommand{\arraystretch}{1.3}
\begin{tabular}{L{0.27\linewidth}L{0.29\linewidth}L{0.33\linewidth}}
\toprule
Analysis & Model or configuration & Role \\
\midrule
Main grounding comparison & Kimi K2.6; Qwen3.6-27B & PathCowork-1T and PathCowork-27B; LoRA rank 32, peak learning rate $10^{-4}$, batch size 16, two epochs; final checkpoint \\
Dermatopathology skill & PathCowork-1T (Kimi K2.6) & Adapted concept localizer evaluated in \textbf{Figure~2e,f} and used by the agent in \textbf{Figure~3} \\
Brain agent comparison & PathCowork v4.4 High; Codex GPT-5.6-sol & Same 30 cases; Claude Code also evaluated, with two missing reports counted as incorrect \\
PathCowork-general & Qwen3.5-4B; LoRA rank 32 & Supervised general navigator; 29 decision slots (Note~\ref{supp:so1}) \\
Small-model development & Qwen3.5-4B; LoRA rank 8 & SFT and subsequent appearance-aware RL experiment (Note~\ref{supp:training}) \\
Glioma format and RL recipe & Qwen3.5-4B; LoRA rank 32 & Prompt-format selection and base versus SFT-initialized RL (Note~\ref{supp:recipe}) \\
Prostate data scaling & Qwen3.5-27B; two epochs & Increasing training-set size and concept-level evaluation (Note~\ref{supp:scaling}) \\
Model-size comparison & Qwen3.5-4B; Qwen3.6-35B-A3B; Qwen3.5-27B; Kimi K2.6 & DEJ localization versus total and active parameter counts \\
Quantization comparison & 27B skin localizer; bf16, 8-bit, 4-bit & Localization, model footprint and latency \\
Training-duration comparison & Kimi K2.6; one or two epochs & Supervised training-loss traces \\
Question-source comparison & Claude / GPT; Haiku 4.5, Sonnet 5, Opus 4.8 & Image-input and separate three-model source comparisons in \textbf{Figure~5b} \\
Benchmark categories & Opus 4.8; Haiku 4.5; GPT-5.5; GPT-5.6-sol; GPT-5.4-mini; Gemini 3.1 & Four categories, with one resident reference \\
MIL feature encoders & UNI2-h; Virchow2; CONCH & Frozen tile embeddings for slide-level grading \\
\bottomrule
\end{tabular}
\end{table}

\clearpage
\begin{table}[H]
\centering
\caption{\textbf{Sources of cases used in agent evaluation.} All three agents were assigned the same cases within each task. Evaluation cases were unseen during grounding training and skill construction.}
\label{tab:agent_cohorts}
\small
\renewcommand{\arraystretch}{1.22}
\begin{tabular}{L{0.19\linewidth}L{0.20\linewidth}L{0.24\linewidth}L{0.25\linewidth}}
\toprule
Task & Source & Evaluation set & Clinical endpoint \\
\midrule
Skin & HISTAI~\cite{nechaev2025histai} & 20 cases; 69 WSIs & Melanoma versus dysplastic nevus \\
Prostate & PANDA~\cite{bulten2022panda} & 20 slides & Cancer detection and ISUP grade \\
Breast & HISTAI~\cite{nechaev2025histai} & 39 cases; grades 1/2/3: 10/14/15 & Nottingham grade and component scores \\
Brain & TCGA-GBM; TCGA-LGG~\cite{tcga2013pancancer} & 30 cases; 15 per class & GBM versus lower-grade glioma \\
Colorectal lymph nodes & LNCO2~\cite{maras2020lnco2} & 77 slides; 10 positive, 67 negative & Node detection and metastasis classification \\
\bottomrule
\end{tabular}
\end{table}

\clearpage
\begin{table}[H]
\centering
\caption{\textbf{Paired comparisons of CoworkMIL with ABMIL.} Means and mean paired differences in quadratic-weighted $\kappa$ are based on 20 matched training seeds per setting, evaluated on the same 30-slide test set. $N$ is the number of training slides. Two-sided exact Wilcoxon signed-rank tests use the archived seed-level scores, stored to four decimal places; none of these contrasts has a zero difference or tied absolute differences. All 15 comparisons are reported; *$P<0.05$, **$P<0.01$. Positive differences favor CoworkMIL.}
\label{tab:mil_paired}
\small
\renewcommand{\arraystretch}{1.2}
\begin{tabular}{lrrrrr}
\toprule
Encoder & $N$ & CoworkMIL & ABMIL & Difference & $P$ \\
\midrule
UNI2-h & 10 & 0.321 & 0.279 & +0.041 & 0.153646 \\
UNI2-h & 20 & 0.544 & 0.466 & +0.078 & 0.001017$^{**}$ \\
UNI2-h & 30 & 0.620 & 0.552 & +0.067 & 0.029575$^{*}$ \\
UNI2-h & 40 & 0.690 & 0.660 & +0.030 & 0.230513 \\
UNI2-h & 50 & 0.712 & 0.724 & -0.012 & 0.348810 \\
Virchow2 & 10 & 0.376 & 0.261 & +0.115 & 0.010689$^{*}$ \\
Virchow2 & 20 & 0.486 & 0.385 & +0.101 & 0.009436$^{**}$ \\
Virchow2 & 30 & 0.620 & 0.469 & +0.151 & 0.001432$^{**}$ \\
Virchow2 & 40 & 0.731 & 0.611 & +0.120 & 0.001432$^{**}$ \\
Virchow2 & 50 & 0.773 & 0.761 & +0.013 & 0.621513 \\
CONCH & 10 & 0.347 & 0.271 & +0.076 & 0.032768$^{*}$ \\
CONCH & 20 & 0.585 & 0.484 & +0.101 & 0.017181$^{*}$ \\
CONCH & 30 & 0.718 & 0.606 & +0.112 & 0.001209$^{**}$ \\
CONCH & 40 & 0.742 & 0.662 & +0.081 & 0.003153$^{**}$ \\
CONCH & 50 & 0.741 & 0.643 & +0.098 & 0.000168$^{**}$ \\
\bottomrule
\end{tabular}
\end{table}

\clearpage
\begin{table}[H]
\centering
\caption{\textbf{EAGLE and MIL pipelines on matched training subsets and test slides.} All rows use the same 20 subsets of 20 training slides and the same 30-slide PANDA test cohort. Quadratic-weighted $\kappa$ is reported as mean $\pm$ sample standard deviation across subsets; MAE is the mean absolute error in ISUP grade. EAGLE uses frozen CTransPath--CHIEF tile selection~\cite{wang2022ctranspath,wang2024chief}, frozen Virchow2 class-token features, mean pooling and a fitted logistic-regression classifier. Its indicated resolution applies to both tile selection and re-encoding. The MIL rows use frozen Virchow2 class-token plus mean patch-token features from native level-0 tiles and a trained attention aggregator; CoworkMIL additionally uses grounding-derived training supervision.}
\label{tab:eagle_matched}
\small
\renewcommand{\arraystretch}{1.25}
\begin{tabular}{lrr}
\toprule
Pipeline & $\kappa$ (mean $\pm$ s.d.) & Mean MAE \\
\midrule
EAGLE, 0.5\,$\mu$m per pixel & $0.430\pm0.160$ & 1.225 \\
EAGLE, 2.0\,$\mu$m per pixel & $0.435\pm0.108$ & 1.298 \\
ABMIL, Virchow2 & $0.385\pm0.133$ & 1.347 \\
CoworkMIL, Virchow2 & $0.486\pm0.120$ & 1.178 \\
\bottomrule
\end{tabular}
\end{table}

\clearpage
\begin{table}[H]
\centering
\caption{\textbf{Composition of the retained recorded-review question bank.} Counts refer to unique retained question identifiers in the two source-bank exports. Environmental sensing comprises two subtypes. The separate 100-question source comparisons and 148-question paired image-input analysis use the subsets described in Methods.}
\label{tab:benchmark_composition}
\small
\renewcommand{\arraystretch}{1.25}
\begin{tabular}{L{0.40\linewidth}L{0.36\linewidth}r}
\toprule
Category & Content & Questions \\
\midrule
Rejection of an inapplicable premise & Finding or premise verification & 128 \\
Multi-slide reasoning & Evidence across fields or slides & 62 \\
Goal-directed navigation & Next region to inspect & 59 \\
Environmental sensing & Viewing scale & 63 \\
 & Tissue location & 59 \\
\midrule
Total & & 371 \\
\bottomrule
\end{tabular}
\end{table}
\clearpage
\section*{Supplementary Video}
\pdfbookmark[\suppchildbookmarklevel]{Supplementary Video 3}{supp.video.3}
\phantomsection\label{video:general}
\noindent\href{https://github.com/zhihuanglab/PathCowork/blob/main/videos/03_Supplementary_Video_3_General_Navigation.mp4}{\textbf{Supplementary Video 3 | General navigation learned from pathologist behavior.}}

Demonstration of PathCowork-general navigating a CAMELYON17 lymph-node slide using a learned viewing policy. The main panel shows the current field and candidate grid cells; the overview shows the viewing trajectory and spatial context. The navigator successively zooms into selected fields, inspects tissue at higher magnification and returns to a broader view before visiting another region. The status panel reports actions, model outputs and measured inference times. Blue overlays indicate reference lesion annotations (filled where read), yellow lines indicate the viewing trajectory and the red box indicates the current field. This recorded example contains 10 steps, with 3.41\,s total model inference time and 3.68\,s recorded runtime. The supplied visualization lasts 17.9\,s, including display holds; its playback timing is preserved. The submission video adds opening and closing title cards and has no audio.
 \restoregeometry
 
\end{document}